\documentclass[article]{elsarticle}
\usepackage{graphics}
\usepackage{graphicx}
\usepackage{epstopdf}
\usepackage{rotating}
\usepackage{pgfplotstable}
\usepackage{natbib}
\usepackage{booktabs}
\usepackage{multirow}

\usepackage{natbib}
\usepackage[latin9]{inputenc}
\usepackage{array}
\usepackage{float}
\usepackage{calc}
\usepackage{textcomp}
\usepackage{multirow}
\usepackage{amsmath}
\usepackage{amssymb}
\usepackage{xcolor}
\usepackage{colortbl}

\floatstyle{ruled}
\newfloat{algorithm}{tbp}{loa}
\providecommand{\algorithmname}{Algorithm}
\floatname{algorithm}{\protect\algorithmname}

\usepackage{lscape}
\usepackage{framed}

\usepackage{latexsym}

\usepackage{algorithm}
\usepackage{algpseudocode}
\usepackage{hyperref}
\usepackage{subcaption}

\usepackage{etoolbox}
\makeatletter
\patchcmd{\ps@pprintTitle}
{Preprint submitted to}
{Preprint accepted in 
	}
{}{}
\makeatother

\usepackage{url}
\definecolor{newcolor}{rgb}{.8,.349,.1}

\journal{Neurocomputing}

\begin{document}

\begin{frontmatter}
	
\title{Label Noise Filtering Techniques to Improve Monotonic Classification}

\author[jaen]{Jos\'e-Ram\'on Cano}
\ead{jrcano@ujaen.es}

\author[ccia]{Juli\'{a}n Luengo}
\ead{julianlm@decsai.ugr.es}

\author[ccia]{Salvador Garc\'{i}a\corref{cor1}}
\ead{salvagl@decsai.ugr.es}

\address[jaen]{Dept. of Computer Science, EPS of Linares, University of Ja\'en, Campus Cient\'ifico Tecnol\'ogico de Linares,  Cintur\'on Sur S/N, Linares 23700, Ja\'en, Spain}
\address[ccia]{Dept. of Computer Science and Artificial Intelligence. University of Granada, 18071 Granada, Spain}

\cortext[cor1]{Corresponding author.}

\begin{abstract}

The monotonic ordinal classification has increased the interest of researchers and practitioners within machine learning community in the last years. In real applications, the problems with monotonicity constraints are very frequent. To construct predictive monotone models from those problems, many classifiers require as input a data set satisfying the monotonicity relationships among all samples. Changing the class labels of the data set (relabelling) is useful for this. Relabelling is assumed to be an important building block for the construction of monotone classifiers and it is proved that it can improve the predictive performance.

In this paper, we will address the construction of monotone datasets considering as noise the cases that do not meet the monotonicity restrictions. For the first time in the specialized literature, we propose the use of noise filtering algorithms in a preprocessing stage with a double goal: to increase both the monotonicity index of the models and the accuracy of the predictions for different monotonic classifiers. The experiments are performed over 12 datasets coming from classification and regression problems and show that our scheme improves the prediction capabilities of the monotonic classifiers instead of being applied to original and relabeled datasets. In addition, we have included the analysis of noise filtering process in the particular case of wine quality classification to understand its effect in the predictive models generated.
\end{abstract}


\end{frontmatter}

\section{Introduction}

In the last years learning with ordinal data has increased the attention of  machine learning research community. Ordinal datasets are characterized by the presence of an ordinal output variable. We find many examples of ordinal data in real life  \citep{Pinto2014,cruz2014metrics,Gutierrez2016}.

The classification with monotonicity constraints, also known as monotonic classification \citep{bendavid89OLM} or isotonic classification \citep{malar13}, is an ordinal classification problem \citep{antoniuk2016v} where a monotonic restriction is present. In monotonic classification, a higher value of an attribute in an example, fixing other values, should not decrease its class assignment. The monotonicity of relations between the dependent and explanatory variables is very usual as a prior knowledge form in data classification \citep{kotlowski13Monot}. To illustrate, while considering a credit card application \citep{ChihChuan14OrdinalSVOCreditRating}, a \$1000 to \$2000 income may be considered a medium value of income in a data set. If a customer $A$ has a medium income, a customer $B$ has a low income (i.e. less than \$1000) and the rest of input attributes remain the same, there is a relationship of partial order between $A$ and $B$: $B < A$. Considering that the application estimates lending quantities as output class, it is quite obvious that the loan that the system should give to customer $B$ cannot be greater than the given to customer $A$. If so, a monotonicity constraint is violated in the decision. Monotonicity is modelling or representing the business logic of the bank's process, which depends on this order.

Several monotonic classification approaches have been proposed in the specialized literature. They include classification trees and rule induction \citep{bendavid95otherprop,potharst00ordtree,caovan03DecTrees,kotlowski09,marsala15,alcala-fdez17}, neural networks \citep{daniels10,zhu2017monotonic},  instance-based learning \citep{bendavid89OLM,Lievens08OSDL,duivesteijn08mKNN} and hybridizations \cite{garcia2016hyperrectangles,garcia2017mongel}. Some of them require the training set to be purely monotone to work correctly, such as the MKNN classifier \cite{duivesteijn08mKNN}. MKNN needs as input a completely monotonic data set and if not, it must be converted in monotonic using relabelling preprocessing methods \cite{duivesteijn08mKNN}. Other classifiers are more permissive allowing as input non-monotonic datasets, but they may not guarantee to make monotonic predictions. An example of this could be the MID algorithm \cite{bendavid95otherprop}. The natural approach is to deal with non-monotonic datasets, as we will justify next. The proposals we find in the literature are designed to work in one way or another or even both at the same time. But sometimes you have to delve into the papers to find out since there is not a catalogue of methods categorized by each type of problem.

Nevertheless, real-world datasets may surely contain noise, which is defined as anything that obscures the relationship between the features of an instance and its class \citep{frenay14,saez2014analyzing}. Among other consequences, noise can adversely impact the classification performances of induced classifiers. By extrapolating to monotonic classification, noise also alters the monotonicity constraints present in the data.

In order to test the performance of monotonic classifiers, the usual trend is to generate datasets that fulfil the monotonicity constraint. The main argument is that models trained on monotonic datasets often have better predictive performance than models trained on the original data \citep{feelders10}. Monotonic datasets can be created by generating artificial data \citep{potharst09} or by relabelling of real data \citep{duivesteijn08mKNN,rademaker12,milstein14}. 

This paper proposes a different general approach to deal with the construction of monotonic models by any classifier. As an alternative, we will consider the examples that do not fulfil the monotonic constraints as noisy examples. For the first time in the literature, we propose the application of noise filtering algorithms in a preprocessing stage for monotonic ordinal classification.

In this study, four classical noise filtering algorithms have been readjusted to this domain. The algorithms considered are the Edited Nearest neighbour (ENN \citep{wilson72enn}), the Relative neighbourhood Graph Editing (RNGE \citep{Sanchez97RNG}), the Iterative Partition Filtering (IPF \citep{Khoshgoftaar07IPF}) and  the Iterative Noise Filter based on the Fusion of Classifiers (MINFFC \cite{SaezGLH16}). The monotonic version of them is noted as MENN, MRNGE, MIPF and MINFFC, respectively. They will be reformulated to detect the non-monotonic samples, which are considered as noisy samples, following different heuristics and strategies. The remaining samples will be those used as input in well-known monotonic classifiers. Our objective is to analyze their performances in comparison with an optimal relabelling scheme of the original datasets in a large collection of datasets.

The paper is organized as follows. In Section \ref{sec:MonotonicClasification} we present the  monotonic classification problem, its formal definition and the monotonic classifiers used in the study. Section \ref{sec:noisefileringalgorithms} is devoted to describing the noise filtering algorithms and their adaptation to satisfy the monotonicity constraints. Section \ref{sec:experimentalFramework} describes the experimental framework. Section \ref{sec:results} analyzes  the results obtained in the empirical study and presents a discussion and analysis of them. In Section \ref{sec:CasesOfStudy} we apply the filter process analyzed in the Winequality-red data set to study its behaviour. Finally, Section \ref{sec:Conclusions} concludes the paper.

\section{Monotonic classification}
\label{sec:MonotonicClasification}

In this section, we introduce the monotonic classification problem and the monotonic classifiers considered in this paper.

\subsection{Problem definition}
\label{sec:problemdefinition}

The property of monotonicity commonly appears in domains of our lives such as natural sciences, natural language, game theory or economics \citep{kotlowski13Monot,tran2015stabilized}. For instance, the case of bankruptcy prediction in companies, where appropriate actions can be taken in time considering the information based on financial indicators taken from their annual reports. The monotonicity is clearly present in the comparison of two companies where one dominates the other on all financial indicators, which supposes that the overall evaluation of the second cannot be higher than the evaluation of the first. This rule could be applied to the credit rating strategy used by banks \citep{ChihChuan14OrdinalSVOCreditRating} as well as for the bankruptcy prediction strategy.

The monotonic ordinal data can be defined as following. Let \textit{D} be a data set with \textit{f} ordinal attributes $A_{1}, ..., A_{f}$ and one output class attribute \textit{Y} having \textit{c} possible ordinal values. The data set consists of \textit{n} examples $x_i$. A partial ordering $\preceq$ on \textit{D}
is defined as

\begin{equation}
	x \preceq x'  \Leftrightarrow A_{j}(x) \leq A_{j}(x'), \forall j=1,...,f
\end{equation}

Two examples \textit{x} and \textit{x'} in space \textit{D} are \textit{comparable} if
either $x \preceq x' $ or $x' \preceq x$, otherwise \textit{x} and \textit{x'} are
\textit{incomparable}. Two examples \textit{x} and \textit{x'} are \textit{identical} if \textit{x} = \textit{x'} and \textit{non-identical} if $x \neq x'$;

Considering this notation, we denote a pair of comparable examples (\textit{x},\textit{x'}) \textit{monotone} if

\begin{equation}
	x\preceq x' \wedge x \neq x'  \wedge Y(x) \leq Y(x')
\end{equation}

\noindent or

\begin{equation}
	x = x' \wedge  Y(x) = Y(x')
\end{equation}

A data set $D$ with \textit{n} examples is monotone if all possible pairs of examples are either monotone or incomparable.

\subsection{Monotonic classifiers}
\label{sec:monotonicclassifiers}

This section reviews the monotone classifiers considered in this paper. They are well-known in the monotonic classification topic.

\begin{itemize}
	
	\item Monotonic $k$-Nearest neighbour Classifier (M$k$NN), proposed in \cite{duivesteijn08mKNN}, can be described as follows.
	
	The first step in this proposal is to relabel the training data aiming to remove all monotonicity violations \cite{feelders2006two}, using as few label changes as possible. 
	
		Starting from the original nearest neighbour rule, the class label assigned to a new data point $x_0$ must lie in the interval $[y_{min},y_{max}]$ in order to preserve the monotonicity conditions, where

	\begin{equation}
		y_{min} = max  \{ y|(x,y) \in D \wedge x\preceq x_0 \}
	\end{equation}
	
		and
	
		\begin{equation}
		y_{max} = min \{  y|(x,y) \in D \wedge x_0 \preceq x  \}
	\end{equation}
	
		$y_{max}$ is the minimum class value of all those instances ($x$) in the data set ($D$) whose attribute values are all bigger than, or equal to, those of $x_0$. And $y_{min}$ is determined as the maximum class value of all those instances ($x$) in the data set ($D$) whose attributes values are all smaller than, or equal to, those of $x_0$.
	 	The nearest neighbour rule considered takes the $k$ nearest neighbours of $x_0$  from $D$ whose labels are included in $[y_{min},y_{max}]$ and makes predictions according to majority voting. If there is not any neighbour which has its class in the range   $[y_{min},y_{max}]$ means that any class assigned will result in a violation of monotonicity and a random class is assigned as prediction \citep{duivesteijn08mKNN}.
	
	\item Ordinal Learning Model (OLM). It was the first method proposed for ordinal classification with monotonicity constraints \citep{bendavid89OLM}. Its goal is to choose a subset $D' \subseteq D$ of training samples in which all meet the monotonicity conditions. The classification of new objects is done by the following function:
	
	\begin{equation}
		f_{OLM}(x) = max \{  y_i: x_i \in D', x_i \preceq x  \}
	\end{equation}
	
	If there is no object from $D'$ which is dominated by $x$, then a class label is assigned by the nearest neighbour rule. $D'$ is chosen to be consistent and not to contain redundant examples. An object $x_i$ is redundant in $D'$ if there is another object $x_j$ such that $x_i \succeq  x_j$ and $y_i = y_j$.
	
	\item Ordinal Stochastic Dominance Learner (OSDL). It was presented in \cite{Lievens08OSDL} and \cite{lievens10} as an instance-based method for ordinal classification with monotonicity constraints based on the concept of ordinal stochastic dominance. The rationale behind it can be given through an example: In life insurance, one may expect a stochastically greater risk to the insurer from older and sicker applicants than from younger and healthier ones. Higher premiums should reflect greater risks and vice versa. There are several definitions of stochastic ordering. The most commonly used is the following: For each example $x_i$, the OSDL computes two mapping functions: one that is based on the examples that are stochastically dominated by $x_i$ with the maximum label (of that subset), and the second is based on the examples that cover (i.e., dominate) $x_i$, with the smallest label. Later, an interpolation between the two class values (based on their position) is returned as a class.
	
	\item Monotone Induction of Decision trees (MID) \citep{bendavid95otherprop}. Its main idea was to modify the conditional entropy in the ID3 algorithm (\cite{Quinlan86ID3}), by adding a term called \emph{order-ambiguity-score}, which adapts the splitting strategy towards the building of monotonic trees. However, this strategy did not guarantee the tree to be a completely monotone function.
\end{itemize}

\section{Label noise filtering in monotonic classification}
\label{sec:noisefileringalgorithms}

As in standard classification, the correct labelling of the training set is crucial to obtain accurate models that will correctly predict new examples.
Most classification algorithms assume that the labelling in the data is correct and follows the underlying distribution without any disturbances.
However, in the real world, this assumption is naive. 
Real-world data is far from being perfect and corruptions usually affect the dependence between the input and output attributes~\cite{frenay14}.
These disturbances will alter or bias the models, hindering their quality.

In classification, the noise may affect the data registration of the input attributes or the labelling process made automatically or by an expert~\cite{Zhu04classnoisevsattributenoise}.
If the noise has affected the input attributes, it is usually named \emph{attribute noise}.
On the other hand, \emph{class noise} or \emph{label noise} means that the noise has corrupted the correct label of some instances.
Some studies have analyzed the impact of these types of noise, indicating that class noise is more harmful than attribute noise, as the bias introduced is greater.
For this reason, in this work, we will focus on label noise and the different ways to tackle its greater impact.

To address this, a conventional strategy consists of \emph{relabelling} the input data to fulfil the monotonicity assumptions \cite{duivesteijn08mKNN,feelders10,rademaker12}. In summary, this process alters the class of those pairs of instances which violate the monotonic restrictions, trying to minimize the total number of changes.

As an alternative strategy, we focus our attention on the approaches to deal with class noise in the specialized literature for classification, which is often grouped into three families.
\begin{enumerate}
	\item We can adapt the classifiers to take into account noise and become \emph{robust learners}, less influenced by noise. Some classifiers were designed as robust against noise from their very conception, as Kernel Logistic Regression~\cite{bootkrajang2014learning}, a robust SVM~\cite{ghosh2015making} or robust Fisher discriminant analysis learners~\cite{lawrence2001estimating,bouveyron2009robust} just to mention a few. We can find adaptations of noise sensitive learners as AdaBoost~\cite{dietterich2000experimental} to become robust learners in recent proposals~\cite{miao2016rboost}.
	
	\item While robust learners can deal with noisy data directly, not all classifiers have been modified to be robust. The solutions designed to make a classifier robust cannot be easily extrapolated to other learners. Thus, a popular option is to apply \emph{noise filtering} methods~\cite{Brodley99identifyingmislabeled, Khoshgoftaar07ImprovingSoftware, Verbaeten03EnsembleMethods}, which work at data level before the classifier is applied. 
	Noise filters aim to detect and eliminate the corrupted instances that would hinder the model built by the classifier, enabling any classifier model to work with noisy datasets.
	
	\item Noise filters are a popular option due to their easy application and independence with the classifier. 
	However, the cost of eliminating instances cannot be disregarded, especially in highly noisy problems. 
	In these cases, the number of instances eliminated would be high enough to produce a data shift in the class borders.
	An optimal preprocessing technique would recover the noisy instances, relabelling them with their true label.
	This family of techniques are known as \emph{data correcting methods}~\cite{Teng99CorrectingNoisy,nicholson2016label}.
\end{enumerate}

Fr\'{e}nay and Verleysen \cite{frenay14} point out that filtering noisy instances is more efficient than correcting them~\cite{CuendetHS07,MirandaGCL09}.
Since correction is never perfect (as a perfect classifier is rarely held), errors will be further added and can accumulate with the noise we intend to remove.
Obtaining correctors with low wrong correction rates are computationally expensive (``less efficient'' as Fr\'{e}nay and Verleysen meant).
Therefore, correcting methods have drawn less attention in the literature than filtering approaches.
Thus, we will focus on filtering approaches for noise.
Among filters, those based on similarity measures and multi-classifiers of ensembles are very popular.

We propose the application of data preprocessing techniques to the original data, which have been successfully applied in the past in similar domains \citep{garcia16,garcia12PAMI,cano08a,cano08b,Han15ExtreLearnMach}. 
In particular, we consider readjusted noise filtering algorithms to tackle the monotonic domain \citep{frenay14}. 
These methods identify and remove some of the examples belonging to the data set presenting a negative effect to the fulfilment of the monotonicity constraints, keeping unaltered the rest of the information held in the original data set. 
Our selection of popular noise filters is justified based on those that obtained the best performances in other learning domains, such as standard classification \citep{SaezGLH16}, imbalanced classification \citep{saez15} or semi-supervised classification \citep{Triguero14NoiseFilters}. 
Next, we describe the filters used in this paper.

\subsection{Monotonic Edited Nearest neighbour (MENN)}
\label{sec:menn}

The Monotonic Edited Nearest neighbour (MENN) evolves from the classical Edited Nearest neighbour algorithm \citep{wilson72enn}. 
It iterates over each instance $x$ in the dataset, and finds the $k$ monotonic nearest neighbours for $x$.
Once such neighbours are found, the $k$ labels are counted.
The label with the highest frequency among the $k$ neighbours' labels is compared to the actual label of $x$.
If the label of $x$ is different than the most frequent label of its $k$ monotonic nearest neighbours,  $x$ is removed from the training set.

The use of the monotonic \textit{k}-nearest neighbours (see Section \ref{sec:monotonicclassifiers} for further details on how to obtain the $k$ monotonic neighbours) instead of the classical \textit{k}-nearest neighbours rule constitutes the adaptation of this algorithm to the monotonic scope. The pseudo-code of MENN is presented in Algorithm \ref{alg:menn}.

\begin{algorithm}[ht]
	\caption{MENN algorithm.}
	\label{alg:menn}
	\begin{algorithmic}
		\Function{MENN}{$T$ - training data, $k$ - number of nearest neighbour}
		\State \textbf{initialize:} $S = T$
		\For{all $x \in S$}
		\State $X' = \emptyset$
		\State     $y_{min} = max  \{ class(x') | x' \in T \wedge x'\leq x \}$
		\State    $y_{max} = min \{  class(x') | x' \in T \wedge x \leq x'  \}$
		\For{$i=1$ to $k$}
		\State Find $ x'_{i} \in T$ s.t. $x \neq x'_{i}$ and $||x-x'_{i}|| = \min_{x^j\in (T \setminus X')}{||x-x^j||}$ and $ class(x'_{i}) \in [y_{min},y_{max}]$
		\State $X' = X' \cup \{x'_{i}\}$
		\EndFor
		\If{$class(x) \neq majorityClass(X')$}
		\State $S = S \setminus \{x\}$
		\EndIf
		\EndFor
		\State \Return $S$
		\EndFunction
	\end{algorithmic}

\end{algorithm}

\subsection{Monotonic Relative neighbourhood Graph Editing (MRNGE)}
\label{sec:mrnge}

The Monotonic Relative neighbourhood Graph Editing (MRNGE) works in a decremental fashion like MENN and it is based on proximity graphs \citep{Sanchez97RNG}. 
In a first phase, MRNGE creates a proximity graph, where the closest instances are linked together in such a way that no other closest example is found between them.
	This is the goal of the Proximity Graph function.
	Once the proximity graph has been built, we can rapidly access to the nearest neighbours of any instance.

In the second phase, MRNGE utilizes the graph to remove the examples by looking at is neighbours.
	However, substantial differences can be found with respect to MENN: instead of only counting the direct neighbours of the reference example $x$, MRNGE will access to the neighbours' neighbour.
	Without the proximity graph, this operation would be computationally expensive.

This ``second order neighbourhood'' aims to diminish the influence of small clusters of noisy examples by extending the examined neighbourhood.
	Thus, MRNGE first examines the $k$ monotonic neighbours for the given example $x$ by accessing the proximity graph. 
	Only in the case where the most frequent class label among the $k$ neighbours differs from the label of $x$, we go further and examine the labels of $x$ neighbours' neighbours (please note that MENN would have stopped here, removing $x$). 
	MRNGE will count the labels of this ``second order neighbourhood'' and only in the case of the most frequent label of the neighbour's neighbours is different to the label of $x$ will mean that $x$ is removed from the dataset. Again, the precomputed proximity graph enables the fast gathering of the neighbours of any given instance.

To work with monotonic classification problems, the graph will be created considering the monotonicity constraints (see Section \ref{sec:problemdefinition}). The Algorithm \ref{alg:mrnge} depicts in detail the procedure MRNGE.

\begin{algorithm}[ht]
	\caption{MRNGE algorithm.}
	\label{alg:mrnge}
	\begin{algorithmic}
		\Function{MRNGE}{$T$ - training data, $k$ - number of nearest neighbour}
		\State \textbf{initialize:} $PG=Proximity\_ Graph(T)$, $S = T$
		\For{all $x \in S$ where $x$ is misclassified by its $k$ neighbours in $PG$}
		\State Consider the subgraph, $R$, given by $x$ and all its graph neighbours from its same class
		\State $S = S \setminus \{x\}$ if the graph neighbourhood of $R$ has a majority of neighbours from different class than $x$.
		\EndFor
		\State \Return $S$
		\EndFunction

		\Function{Proximity\_Graph}{$T$ - training data}
		\State \textbf{initialize:} $PG=(V,E)$, $V=T$, $E=\emptyset$
		\For{all $x_i \in E$}
		
		\For{all $x_j \in E$}
		
		\For{all $x_k \in E$}
		
		\If{ ( k $\neq$ i and k $\neq$ j  )}
		\State $d_{ij}$=$EuclideanDistance$($x_i$,$x_j$)
		\State $d_{ik}$=$EuclideanDistance$($x_i$,$x_k$)
		\State $d_{jk}$=$EuclideanDistance$($x_j$,$x_k$)
		\If{ $d_{ij}^2 \leq d_{ik}^2 + d_{jk}^2  $  }
		\State $E= E \bigcup (x_i$,$x_j)$
		\EndIf         	
		\EndIf
		
		\EndFor
		\EndFor
		\EndFor
		\State \Return $PG$
		\EndFunction
	\end{algorithmic}

\end{algorithm}

\subsection{Monotonic Iterative Partition Filtering Editing (MIPF)}
\label{sec:mipf}

The Monotonic Iterative Partition Filtering (MIPF) is a global noise filter which applies a classifier to several subsets of the training data set to detect possible noisy examples.
It removes noisy instances in multiple iterations until the number of identified noisy examples, for a number of consecutive iterations, is less than a percentage of the size of the original training data set \citep{Khoshgoftaar07IPF}. The classifier embedded in the classic Iterative Partition Filtering  algorithm is the C4.5 \citep{QuinlanC45}. 
 Since C4.5 does not takes into account the monotonic restrictions of examples, for our MIPF proposal, we consider the ordinal interpretation of C4.5 \citep{frank01otherprop} as the base classifier. It is worth mentioning that the ordinal C4.5 does not produce monotonic models but ensures ordinal classification. MIPF is described in Algorithm \ref{alg:mipf}.

\begin{algorithm}[ht]
	\caption{MIPF algorithm.}
	\label{alg:mipf}

	\begin{algorithmic}
		\Function{MIPF}{$T$ - dataset with Monotonic Violations, $\Gamma$ - number of subsets, $y$ -  amount of good data to be eliminated in each step, $p$ - minimum percentage of noisy instances to continue}
		\State \textbf{initialize:} $T_G = \{\}$, $F=$ Ordinal C4.5
		\Repeat
		\State Split the training data set $T$ into $T_i, i = 1 \ldots \Gamma$ equal sized subsets		
		\For{each subset $T_i$}
		\State Use $\{T_j, j \neq i\}$ to train $F$ resulting in $F^i$ different classifiers
		\EndFor
		\State $D_N=\{\}, D_G=\{\}$
		\For{each instance $t$ in $T$}
		\State Classify $t$ with every $F^i$
		\If{$t$ is voted as noisy}
		\State $D_N = D_N \cup t$
		\EndIf
		\EndFor
		\State $D_G = \{ t_l \in T | t_l \notin D_N; l=1,...,y \} $
		\State $T_G = T_G \cup D_G$
		\State $T= T - \{D_N \cup D_G \}$

		\Until{$|D_N| < p \cdot |T|$}
		\State \Return $T \cup T_G$
		\EndFunction
	\end{algorithmic}
	
\end{algorithm}

\subsection{Monotonic Iterative class noise filter based on the fusion of classifiers  (MINFFC)}
\label{sec:minffc}

Finally, we include the Monotonic Iterative Noise Filter based on the Fusion of Classifiers (MINFFC), based on the proposal for standard classification made in~\cite{SaezGLH16}.
It is an advanced filter that gathers strategies and techniques from different kinds of noise filters: ensembles of filters, noise scoring and an iterative processing.
It is based on an ensemble of classifiers to select a candidate set of noisy instances.
It first performs a preliminary filtering, generating a temporarily reduced set, which is then used to generate the final filtering ensemble.
The final filtering ensemble only points out the set of noisy candidate instances.
Such a set is the foundation to compute a noise score for all the instances, thus removing those that are detected as noise with a score greater than 0.
The process is applied iteratively until the number of removed examples is below a percentage for a given number of iterations.

For monotonic classification, the computation of the \emph{noiseScore} is adapted from that described in ~\cite{SaezGLH16}.
We multiply the \emph{clean} value of each instance by the NMI1 index (see Section~\ref{sec:metrics}) previously computed.
MINFFC is described in Algorithm \ref{alg:minffc}.

\begin{algorithm}[p]
	\caption{MINFFC algorithm.}
	\label{alg:minffc}
	\begin{algorithmic}
		\Function{MINFFC}{$T$ - dataset with Monotonic Violations, 
			$g$ - number of iterations to stop, $p$ - minimum percentage of noisy instances to stop, $k$ - number of neighbours to compute the noise score, $NMI1$ - Non-monotonic index for each instance}
		\State \textbf{initialize:} $T_G = \{\}$, $F_1=$ Ordinal C4.5, $F_2=$ 3-NN, $F_3=$ Logistic Regression, $it=0$
		\Repeat
		\State Split the training data set $T$ into $T_i, i = 1 \ldots \Gamma$ equal sized subsets		
		\For{each subset $T_i$}
		\State Use $\{T_j, j \neq i\}$ to train $F^i_j, j = 1 \ldots 3$ different classifiers
		\EndFor
		\State $D_N=\{\}, D_G=\{\}$
		\For{each instance $t$ in $T$}
		\State Classify $t$ with every $F^i_j$ where $t \notin T_i$
		\If{$t$ is voted as noisy}
		\State $D_N = D_N \cup t$
		\EndIf
		\EndFor
		\State $D_G = \{ t_l \in T | t_l \notin D_N; l=1,...,y \} $
		\State Split the training data set $D_G$ into $D_i, i = 1 \ldots \Gamma$ equal sized subsets		
		\For{each subset $D_i$}
		\State Use $\{D_j, j \neq i\}$ to train $F^i_j, j = 1 \ldots 3$ different classifiers
		\EndFor
		\State $D'_N=\{\}$
		\For{each instance $t$ in $T$}
		\State Classify $t$ with every $F^i_j$ where $t \notin T_i$
		\If{$t$ is voted as noisy}
		\State $D'_N = D'_N \cup t$
		\EndIf
		\EndFor
		\For{each instance $t$ in $T$}
		\State score = \emph{noiseScore}($t$) $\cdot NMI1(t)$
		\If{score $\geq$ 0}
		\State $T = T - t$
		\EndIf
		\EndFor
		\If{$|D_N| < p \cdot |T|$}
		\State $it = it -1$
		\Else
		\State $it = g$
		\EndIf
		\Until{it = 0}
		\State \Return $T$
		\EndFunction
	\end{algorithmic} 

\end{algorithm}

\section{Experimental framework}
\label{sec:experimentalFramework}

In this section, we present the experimental framework developed to analyze the proposal of application of four well-known noise filtering algorithms readjusted to work in this domain. 
Section \ref{sec:datasets} introduces the datasets used in the comparison.
Section \ref{sec:metrics} describes the metrics used to evaluate the compared methods.
Section \ref{sec:parameters} lists all the parameters used for each method in the experimental comparison. 
Finally, Section \ref{sec:statistical_procedures} describes the statistical procedures employed to support the analysis carried out.
	
\subsection{datasets}
\label{sec:datasets}

The study includes 12 datasets whose class attribute can be expressed as ordinal and presents a monotonic relationship with the features. Four datasets are actual classical ordinal classification datasets commonly used in this field (Era, Esl, Lev and Swd \citep{bendavid89OLM}). 
The other 7 are regression datasets whose class attribute was discretized into 4 categorical values, maintaining the class distribution balanced. 
The last one is Winequality-red, a well-known data set in  classification which is extensively analyzed in Section \ref{sec:CasesOfStudy}.  
All of the 12 datasets are classical problems used in the classification scope and extracted from the UCI \citep{Bache+Lichman:2013} and KEEL\footnote{http://www.keel.es/datasets.php} repositories \citep{alcala2010keel,triguero17}. 

In order to evaluate the performance of the different approaches with different amounts of monotonic violations, we have generated three corrupted versions of each dataset.
These altered versions are created by changing a $noise\%$ of instances by relabelling them with a new class label.
The new label can only be the precedent or the following one, thus generating realistic disorders in terms of monotonic violations:
\begin{itemize}
	\item When corrupting to the next label, the modified instance is not the corrupted datum, but the instances that have been surpassed by such noisy instance.
	\item When applying this corruption scheme, the class order is considered as cyclical. If the last label is to be corrupted to the next class label, we will select the first label as the new output value. On the other hand, if the first label is to be corrupted to the previous label, we will select the last label instead. These exceptional cases will induce even more noise than the intermediate labels.  
\end{itemize}
Such a noise introduction scheme follows the NAR mechanism as described in \cite{frenay14}, in which the true label has influence in the observed (and possibly corrupted) label.
We have applied $noise\% = 10\%, 20\%$ and $30\%$ levels only in the training partitions to simulate from low to high noisy scenarios.
Since the true labels are known, we can later examine the performance of the preprocessing approaches in term of the well and wrongly filtered instances.

All the algorithms are run using run a 10-fold cross validation scheme (10-fcv).
For all the training partitions, three different noisy versions are generated (with different seeds) for each noise level.
Therefore, we obtain 30 executions per dataset and noise level.
	
Table \ref{tab:datasets} shows the names of datasets, their number of instances, attributes, and classes. 
In the last column, we present the number of features which present monotonic relation with the class, using for this the RMI measure \cite{Hu12REMT}. This metric is calculated using each feature and the class and takes values in the range [-1; 1], where a -1 means that the relationship is totally inverse (if the feature increases, the class decreases), and a 1 represents a completely direct relation (if the feature increases, the class increases). When
the RMI value is in the range [-0.1; 0.1], we consider that there is no relation between the feature and the class. Counting features whose RMI value is out of that range, we calculate the number of features with order respect to the class. If all the features present order, we refer that the data set presents total ordering. In case of that number is lower than the total number of features, the data set presents partial order.  
In this paper, instances having missing values have been ignored.	
	
	\begin{table}[t]
		\centering
		\caption{Description of the 12 datasets used in the study.}
		\label{tab:datasets}
		\begin{tabular}{|l|cccc|}
			
			\hline  Data set & Ins.  & At. & Cl. &  \#Mon. Features \\
			\hline
			\hline 
			Balance & 625  & 4   & 3  & 4 \\
			BostonHousing & 506 & 12 & 4 & 10 \\
			Car & 1728  &  6 & 4 & 6\\
			Era &  1000 & 4   &  9  & 4 \\
			Esl  & 488  & 4  &  9 & 4\\
			Lev  & 1000  & 4 & 5  & 4\\
			CPU & 209 & 6 & 4 & 6 \\
			QualitativeBankruptcy & 250 & 6 & 2  & 6\\
			Swd & 1000  & 10 &  4 & 7 \\
			WindsorHousing & 545 & 11 & 2  & 2 \\
			Winequality-red & 1599 & 11 & 11 & 8\\
			Wisconsin  & 683  & 9 & 2 & 9\\
			\hline
			
		\end{tabular}
		
	\end{table}
	
\subsection{Evaluation metrics}
\label{sec:metrics}

In order to compare the four monotonic filters, we will use five metrics commonly employed in the monotonic classification field. 
They are listed as follows:
	\begin{itemize}

		\item Mean Absolute Error (MAE), is calculated as the sum of the absolute values of the errors and then dividing it by the number of classifications. 
		Various studies conclude that MAE is one of the best performance metrics in ordered classification \citep{gaudette09MAE, japkowicz14BookEvaluateLeargAlg}.
		
		\item Accuracy (ACC) is computed as the percentage of correctly classified instances. Is a traditional measure in the classification topic that we include as a reference metric.


		\item Non-Monotonicity Index 1 (NMI1) \citep{Daniels06NMI1}, is defined as the number of clash-pairs divided by the total number of pairs of examples in the data set:
		
		\begin{equation}
			\text{NMI1}= \dfrac{1}{n(n-1)}\sum_{x \in D} \text{NClash}(x)
		\end{equation}
		
		\noindent where $x$ is an example from the data set $D$.  $\text{NClash}(x)$ is the number of examples from $D$ that do not meet the monotonicity restrictions (or clash) with $x$ and $n$ is the number of instances in $D$.
		
		\item Non-Monotonicity Index 2 (NMI2) \citep{milstein14}, is defined as the number of non-monotone examples divided by the total number of examples:
		
		\begin{equation}
			\text{NMI2}= \dfrac{1}{n}\sum_{x \in D} \text{Clash}(x)
		\end{equation}
		
		\noindent where $\text{Clash}(x)$ = 1 if $x$ clashes with some examples in $D$, and 0 otherwise. If $\text{Clash}(x)$ = 1, $x$ is called a non-monotone example.
		
		\item Non-Comparable. This is a metric related to the number of pairs of non-comparable instances in the data set. Two instances $x$ and $x'$ are non-comparable if they do no satisfy $x\preceq x' \wedge x \neq x'$. This measure is also considered due to the fact that for some monotonic classifiers, it is harder to construct accurate models agreeing the number of non-comparable pairs raises.		
		\item \emph{Size} of the subset selected using the noise filtering algorithms. We include it to analyze the noise removing capabilities of each method.

	\end{itemize}

\subsection{Parameters configuration}
\label{sec:parameters}

See Table \ref{tab:parameters}. 
The parameters of the four techniques that are the same as the original algorithms have been fixed following the suggestions of their authors.
These original parameters are related to the neighbourhood size or the number of partitions (i.e. base classifiers used), which have a great impact in the computational cost of the filtering methods.
We understand that the authors of the standard classification versions obtained a compromise between the noise detection accuracy and the computational complexity that can be translated to our monotonic problems.
	
	\begin{table}[ht]
		\centering
		\caption{Parameters considered for the algorithms compared. Underlined parameters has been optimized by grid search. The others have been fixed following the suggestions of the original authors.}
		\resizebox{0.8\textwidth}{!}{
			\begin{tabular}{|l|c|}
				\hline
				Algorithm & Parameters\\
				\hline
				\hline
				MENN & $k = 3$\\
				\hline
				MRNGE & firstOrderEdition = true\\
				\hline
				MIPF & numberPartitions = 5, consensus filter\\
				& \underline{confidence = 0.25, 2 items per leaf}\\
				\hline
				MINFFC & numberPartitions = 3,  majority filter\\
				& $k = 3$, \underline{threshold= 0}\\
				& \underline{confidence = 0.25, 2 items per leaf}\\
				\hline
				\hline
				M$k$NN & $k$ = 3, distance = euclidean\\
				\hline
				OLM & modeResolution = conservative\\
				& modeClassification = conservative\\
				\hline
				OSDL & classificationType = media, balanced = No\\
				& weighted = No, tuneInterpolationParameter = No, \\
				&lowerBound = 0, upperBound = 1\\
				&interpolationParameter = 0.5, interpolationStepSize = 10\\
				\hline
				MID & confidence = 0.25, 2 items per leaf, R = 1\\
				\hline
			\end{tabular}
		}
		\label{tab:parameters}
	\end{table}
	
\subsection{Statistical procedures} 
\label{sec:statistical_procedures}

Several hypothesis testing procedures are considered to determine the most relevant differences found among the methods \citep{sheskin11StatTest}. The use of nonparametric tests is preferred over parametric ones, since the initial conditions that guarantee the reliability of the latter may not be satisfied. 
Friedman statistical test, a multiple comparison tests,  is used to contrast the behaviour of the algorithms \citep{garcia09StatTest} by ranking them and showing which are significantly different than the best thanks to Holm's posthoc test.

\section{Analysis on the usage of monotonic filters to remove non-monotonic instances}
\label{sec:results}

This section is devoted to analyzing the results obtained, providing a summary of results including graphics and statistical outcome. 
We present the results considering two perspectives:

\begin{enumerate}
	\item We compare the behaviour of the algorithms using  prediction quality measures MAE and Accuracy. 
	In addition to the noise removal algorithms, we include the results obtained using the original datasets as input and  datasets after relabelling the training partitions (keeping the tests partitions as they are). As we mentioned before, the relabelling  is introduced to study its behaviour in the real data when new unseen examples have to be classified using as input the relabeled training sets. The relabelling used is the optimal proposal described in \citep{duivesteijn08mKNN}.
	
From the original data, one can create a graph representing the monotonicity violations between the instances. The instances correspond to vertex, and the violations are edges. A subset of the vertexes of a graph is an independent set if there are not two vertices in the subset that are adjacent \cite{rademaker12}. 
In a monotonic violation graph, a maximum weights independent set corresponds to a monotone subset of the maximum size.

Re-labelling the complement of the maximum independent set produces a completely monotonic set, with the fewest number of label changes in the instances. Although finding the maximum independent set is an NP-Hard problem, this is not the case for partial order graphs (comparability graphs or networks). In these graphs, the maximum independent set corresponds to a maximum antichain in the corresponding partial order and can be calculated in polynomial time by resolution of a minimum flow problem on a transportation network that is simply created from the comparability graph \cite{mohring1985algorithmic}. As can be noted, the monotone violation network is a comparability network.
	
	This analysis is carried out in Section~\ref{sec:performance_analysis}.
	
	\item We study the algorithms using different metrics to study how the filtering and relabelling process affects the monotonic properties of the datasets: NMI1, NMI2, \emph{Non-Comparable} and \emph{Size}. Section~\ref{sec:metric_analysis} is devoted to study such measures.
	
\end{enumerate}

\subsection{Performance measures}
\label{sec:performance_analysis}

In this section, we provide the accuracy and MAE results for all the classifiers and preprocessing strategies described above. 
These measures are computed over the test partitions after applying Relabelling, MENN, MRNGE, MIPF and MINFFC.
We also include the absence of preprocessing, named as \emph{No preprocessing}, to show the consequences of leaving an increasing amount of monotonic violations in the training set.

Table~\ref{tab:accuracy} shows the averaged accuracy values for all the datasets.
The algorithm with the best value is stressed in bold.
As we can appreciate, MIPF is the best performing technique on average, except for OSDL at 30\% noise, where MINFFC is the best choice.
It is interesting to note that Relabelling is not performing as well as expected, obtaining poor accuracy values when we introduce higher amounts of noise by forcing violations in the monotonic constraints.

On the right side of Table~\ref{tab:accuracy}  the rankings of Friedman's test are shown. 
Friedman's test rejects the null-hypothesis in all cases.
MIPF is again the best ranked algorithm, except for OSDL when we consider the original datasets.
Relabelling is the worst ranked algorithm when we add any quantity of noise, suggesting that it is not a suitable technique in these environments.
Grey cells depict when Holm's posthoc test indicates that there is a $p$-value below $0.05$ rejecting the null hypothesis in favour of the control algorithm.
The large amounts of shaded cells support the choice of MIPF as the best performing algorithm, while MINFFC is the only alternative that it is not statistically different.

\begin{table}[t]
	\centering
	\caption{Average accuracy results for the filters and relabel with all the classifiers and each noise level. Best values are stressed in bold with respect to full precision results.
	Friedman rankings are provided indicating the best rank (control algorithm) in bold.
	When Holm's posthoc indicates a $p$-value $< 0.05$ the cell of the compared algorithm is grayed, whereas a $p$-value $< 0.1$ is indicated by underling the rank of the compared algorithm.}
	\resizebox{\textwidth}{!}{
	\begin{tabular}{c|l|cccc|cccc}
	    \multicolumn{1}{r}{} &       & \multicolumn{4}{c|}{Accuracy averages} & \multicolumn{4}{c}{Friedman's rankings} \\
		\cmidrule{2-10}    \multicolumn{1}{r|}{} & Preprocessing & 0\% (Original) & 10\%  & 20\%  & 30\%  & 0\% (Original) & 10\%  & 20\%  & 30\% \\
		\midrule
		\multirow{6}[2]{*}{MKNN} & No preprocessing & 0.69  & 0.65  & 0.60  & 0.55  & 3.00  & \cellcolor[rgb]{ .906,  .902,  .902}3.18 & \cellcolor[rgb]{ .906,  .902,  .902}3.55 & \cellcolor[rgb]{ .906,  .902,  .902}3.64 \\
		& Relabelling & 0.52  & 0.43  & 0.40  & 0.37  & \cellcolor[rgb]{ .906,  .902,  .902}5.00 & \cellcolor[rgb]{ .906,  .902,  .902}5.45 & \cellcolor[rgb]{ .906,  .902,  .902}5.36 & \cellcolor[rgb]{ .906,  .902,  .902}5.27 \\
		& MENN & 0.61  & 0.58  & 0.54  & 0.51  & \underline{3.36}  & \cellcolor[rgb]{ .906,  .902,  .902}3.50 & \cellcolor[rgb]{ .906,  .902,  .902}3.50 & \cellcolor[rgb]{ .906,  .902,  .902}3.86 \\
		& MRNGE & 0.51  & 0.47  & 0.44  & 0.43  & \cellcolor[rgb]{ .906,  .902,  .902}4.55 & \cellcolor[rgb]{ .906,  .902,  .902}4.68 & \cellcolor[rgb]{ .906,  .902,  .902}4.59 & \cellcolor[rgb]{ .906,  .902,  .902}4.14 \\
		& MIPF & \textbf{0.71} & \textbf{0.70} & \textbf{0.69} & \textbf{0.65} & \textbf{1.91} & \textbf{1.59} & \textbf{1.55} & \textbf{1.64} \\
		& MINFFC & 0.69  & 0.68  & 0.66  & 0.63  & 3.18  & 2.59  & 2.45  & 2.45 \\
		\midrule
		\multirow{6}[2]{*}{MID} & No preprocessing & 0.72  & 0.69  & 0.65  & 0.61  & 2.45  & 2.68  & 2.82  & 2.73 \\
		& Relabelling & 0.53  & 0.44  & 0.40  & 0.38  & \cellcolor[rgb]{ .906,  .902,  .902}4.73 & \cellcolor[rgb]{ .906,  .902,  .902}5.18 & \cellcolor[rgb]{ .906,  .902,  .902}5.27 & \cellcolor[rgb]{ .906,  .902,  .902}5.27 \\
		& MENN & 0.60  & 0.58  & 0.55  & 0.51  & \cellcolor[rgb]{ .906,  .902,  .902}3.95 & \cellcolor[rgb]{ .906,  .902,  .902}3.77 & \cellcolor[rgb]{ .906,  .902,  .902}3.77 & \cellcolor[rgb]{ .906,  .902,  .902}4.14 \\
		& MRNGE & 0.51  & 0.47  & 0.44  & 0.42  & \cellcolor[rgb]{ .906,  .902,  .902}4.68 & \cellcolor[rgb]{ .906,  .902,  .902}4.68 & \cellcolor[rgb]{ .906,  .902,  .902}4.77 & \cellcolor[rgb]{ .906,  .902,  .902}4.41 \\
		& MIPF & \textbf{0.73} & \textbf{0.71} & \textbf{0.69} & \textbf{0.65} & \textbf{1.91} & \textbf{1.68} & \textbf{1.82} & \textbf{1.91} \\
		& MINFFC & 0.69  & 0.67  & 0.66  & 0.63  & \underline{3.27}  & \underline{3.00}  & 2.55  & 2.55 \\
		\midrule
		\multirow{6}[2]{*}{OLM} & No preprocessing & 0.56  & 0.50  & 0.49  & 0.47  & 3.68  & \cellcolor[rgb]{ .906,  .902,  .902}3.91 & \cellcolor[rgb]{ .906,  .902,  .902}3.82 & \cellcolor[rgb]{ .906,  .902,  .902}3.59 \\
		& Relabelling & 0.49  & 0.41  & 0.37  & 0.35  & \cellcolor[rgb]{ .906,  .902,  .902}4.45 & \cellcolor[rgb]{ .906,  .902,  .902}5.00 & \cellcolor[rgb]{ .906,  .902,  .902}4.91 & \cellcolor[rgb]{ .906,  .902,  .902}4.82 \\
		& MENN & 0.54  & 0.50  & 0.47  & 0.46  & 3.09  & \cellcolor[rgb]{ .906,  .902,  .902}3.32 & \cellcolor[rgb]{ .906,  .902,  .902}3.59 & \cellcolor[rgb]{ .906,  .902,  .902}3.77 \\
		& MRNGE & 0.45  & 0.40  & 0.38  & 0.37  & \cellcolor[rgb]{ .906,  .902,  .902}4.14 & \cellcolor[rgb]{ .906,  .902,  .902}4.68 & \cellcolor[rgb]{ .906,  .902,  .902}4.41 & \cellcolor[rgb]{ .906,  .902,  .902}4.68 \\
		& MIPF & \textbf{0.60} & \textbf{0.59} & \textbf{0.57} & \textbf{0.55} & \textbf{2.50} & \textbf{1.68} & \textbf{1.73} & \textbf{1.68} \\
		& MINFFC & 0.58  & 0.58  & 0.56  & 0.55  & 3.14  & 2.41  & 2.55  & 2.45 \\
		\midrule
		\multirow{6}[2]{*}{OSDL} & No preprocessing & 0.59  & 0.46  & 0.44  & 0.40  & \textbf{2.64} & \cellcolor[rgb]{ .906,  .902,  .902}3.77 & \underline{3.55}  & \cellcolor[rgb]{ .906,  .902,  .902}3.91 \\
		& Relabelling & 0.49  & 0.39  & 0.37  & 0.35  & \cellcolor[rgb]{ .906,  .902,  .902}4.45 & \cellcolor[rgb]{ .906,  .902,  .902}4.64 & \cellcolor[rgb]{ .906,  .902,  .902}4.73 & \cellcolor[rgb]{ .906,  .902,  .902}4.36 \\
		& MENN & 0.58  & 0.52  & 0.49  & 0.47  & 3.41  & \cellcolor[rgb]{ .906,  .902,  .902}3.55 & \underline{3.41}  & \cellcolor[rgb]{ .906,  .902,  .902}3.77 \\
		& MRNGE & 0.51  & 0.45  & 0.42  & 0.42  & 3.68  & \cellcolor[rgb]{ .906,  .902,  .902}4.05 & \cellcolor[rgb]{ .906,  .902,  .902}4.32 & \cellcolor[rgb]{ .906,  .902,  .902}4.05 \\
		& MIPF & \textbf{0.59} & \textbf{0.58} & \textbf{0.56} & 0.54  & 3.00  & \textbf{1.86} & \textbf{2.00} & \textbf{2.09} \\
		& MINFFC & 0.57  & 0.57  & 0.56  & \textbf{0.54} & 3.82  & 3.14  & 3.00  & 2.82 \\
		\bottomrule
	\end{tabular}}
	\label{tab:accuracy}%
\end{table}%

Next, we include the averaged MAE results in Table~\ref{tab:mae} with the same format as that described for Table~\ref{tab:accuracy}.
In this case, MIPF is always the algorithm with the best average MAE, as well as the most robust choice as its MAE increases less than the other techniques as noise increases.
MRNGE and Relabelling are the algorithms with the worst  average MAE, being greatly affected by noise.

If we attend to the Friedman's test rankings and $p$-values, we may indicate that MIPF is the best choice overall.
Again, Friedman's test rejects the null-hypothesis in all cases.
Only not preprocessing for MID and OSDL and MINFFC in some cases are comparable to MIPF, as no statistical differences are found.
However, the better performance of MIPF for all the four classifiers promotes it as the most recommendable strategy to apply in noisy environments for monotonic classification.

\begin{table}[t]
	\centering
	\caption{Average MAE results for the filters and relabelling with all the classifiers and each noise level. Best values are stressed in bold with respect to full precision results.
		Friedman rankings are provided indicating the best rank (control algorithm) in bold.
		When Holm's posthoc indicates a $p$-value $< 0.05$ the cell of the compared algorithm is greyed, whereas a $p$-value $< 0.1$ is indicated by underling the rank of the compared algorithm.}
	\resizebox{\textwidth}{!}{
	\begin{tabular}{c|l|cccc|cccc}
		  \multicolumn{1}{r}{} &       & \multicolumn{4}{c|}{MAE averages} & \multicolumn{4}{c}{Friedman's rankings} \\
		\cmidrule{2-10}    \multicolumn{1}{r|}{} & Preprocessing & 0\% (Original) & 10\%  & 20\%  & 30\%  & 0\% (Original) & 10\%  & 20\%  & 30\% \\
		\midrule
		\multirow{6}[2]{*}{MKNN} & No preprocessing & 0.44  & 0.48  & 0.54  & 0.60  & \underline{3.27}  & \cellcolor[rgb]{ .906,  .902,  .902}3.36 & \cellcolor[rgb]{ .906,  .902,  .902}3.55 & \cellcolor[rgb]{ .906,  .902,  .902}3.91 \\
		& Relabelling & 0.75  & 0.86  & 0.93  & 0.97  & \cellcolor[rgb]{ .906,  .902,  .902}4.73 & \cellcolor[rgb]{ .906,  .902,  .902}5.36 & \cellcolor[rgb]{ .906,  .902,  .902}5.27 & \cellcolor[rgb]{ .906,  .902,  .902}5.09 \\
		& MENN & 0.52  & 0.55  & 0.59  & 0.62  & 3.09  & \cellcolor[rgb]{ .906,  .902,  .902}3.23 & \underline{2.95}  & \cellcolor[rgb]{ .906,  .902,  .902}3.32 \\
		& MRNGE & 0.89  & 0.97  & 1.00  & 1.03  & \cellcolor[rgb]{ .906,  .902,  .902}4.73 & \cellcolor[rgb]{ .906,  .902,  .902}4.86 & \cellcolor[rgb]{ .906,  .902,  .902}4.86 & \cellcolor[rgb]{ .906,  .902,  .902}4.59 \\
		& MIPF & \textbf{0.38} & \textbf{0.40} & \textbf{0.41} & \textbf{0.44} & \textbf{1.91} & \textbf{1.41} & \textbf{1.64} & \textbf{1.64} \\
		& MINFFC & 0.45  & 0.47  & 0.50  & 0.52  & \underline{3.27}  & \underline{2.77}  & 2.73  & 2.45 \\
		\midrule
		\multirow{6}[2]{*}{MID} & No preprocessing & 0.36  & 0.39  & 0.44  & 0.48  & 2.09  & 2.64  & 2.73  & 2.55 \\
		& Relabelling & 0.74  & 0.84  & 0.92  & 0.95  & \cellcolor[rgb]{ .906,  .902,  .902}4.86 & \cellcolor[rgb]{ .906,  .902,  .902}5.36 & \cellcolor[rgb]{ .906,  .902,  .902}5.27 & \cellcolor[rgb]{ .906,  .902,  .902}5.18 \\
		& MENN & 0.52  & 0.54  & 0.57  & 0.61  & \cellcolor[rgb]{ .906,  .902,  .902}3.82 & \cellcolor[rgb]{ .906,  .902,  .902}3.41 & \cellcolor[rgb]{ .906,  .902,  .902}3.41 & \cellcolor[rgb]{ .906,  .902,  .902}3.86 \\
		& MRNGE & 0.89  & 0.97  & 1.03  & 1.06  & \cellcolor[rgb]{ .906,  .902,  .902}5.05 & \cellcolor[rgb]{ .906,  .902,  .902}5.05 & \cellcolor[rgb]{ .906,  .902,  .902}5.05 & \cellcolor[rgb]{ .906,  .902,  .902}4.95 \\
		& MIPF & \textbf{0.35} & \textbf{0.37} & \textbf{0.40} & \textbf{0.43} & \textbf{1.91} & \textbf{1.36} & \textbf{1.55} & \textbf{1.82} \\
		& MINFFC & 0.46  & 0.49  & 0.52  & 0.54  & \underline{3.27}  & \cellcolor[rgb]{ .906,  .902,  .902}3.18 & \underline{3.00}  & 2.64 \\
		\midrule
		\multirow{6}[2]{*}{OLM} & No preprocessing & 0.68  & 0.76  & 0.77  & 0.80  & 3.68  & \cellcolor[rgb]{ .906,  .902,  .902}3.91 & \underline{3.36}  & \underline{3.32} \\
		& Relabelling & 0.83  & 0.93  & 1.00  & 1.03  & \cellcolor[rgb]{ .906,  .902,  .902}4.18 & \cellcolor[rgb]{ .906,  .902,  .902}4.64 & \cellcolor[rgb]{ .906,  .902,  .902}4.64 & \cellcolor[rgb]{ .906,  .902,  .902}4.91 \\
		& MENN & 0.69  & 0.75  & 0.79  & 0.80  & 3.00  & \underline{3.14}  & \cellcolor[rgb]{ .906,  .902,  .902}3.50 & \cellcolor[rgb]{ .906,  .902,  .902}3.41 \\
		& MRNGE & 1.03  & 1.11  & 1.20  & 1.23  & \cellcolor[rgb]{ .906,  .902,  .902}4.59 & \cellcolor[rgb]{ .906,  .902,  .902}5.05 & \cellcolor[rgb]{ .906,  .902,  .902}4.95 & \cellcolor[rgb]{ .906,  .902,  .902}5.05 \\
		& MIPF & \textbf{0.57} & \textbf{0.57} & \textbf{0.62} & \textbf{0.62} & \textbf{2.41} & \textbf{1.59} & \textbf{1.82} & \textbf{1.77} \\
		& MINFFC & 0.65  & 0.67  & 0.68  & 0.67  & 3.14  & 2.68  & 2.73  & 2.55 \\
		\midrule
		\multirow{6}[2]{*}{OSDL} & No preprocessing & 0.54  & 0.66  & 0.68  & 0.72  & \textbf{2.45} & 3.14  & 3.09  & 3.27 \\
		& Relabelling & 0.80  & 0.91  & 0.98  & 1.00  & \cellcolor[rgb]{ .906,  .902,  .902}4.64 & \cellcolor[rgb]{ .906,  .902,  .902}5.09 & \cellcolor[rgb]{ .906,  .902,  .902}4.73 & \cellcolor[rgb]{ .906,  .902,  .902}4.45 \\
		& MENN & 0.57  & 0.63  & 0.65  & 0.67  & 3.14  & \cellcolor[rgb]{ .906,  .902,  .902}3.45 & \cellcolor[rgb]{ .906,  .902,  .902}3.41 & \cellcolor[rgb]{ .906,  .902,  .902}3.77 \\
		& MRNGE & 0.86  & 0.96  & 1.00  & 1.03  & \underline{3.77}  & \cellcolor[rgb]{ .906,  .902,  .902}4.05 & \cellcolor[rgb]{ .906,  .902,  .902}4.41 & \cellcolor[rgb]{ .906,  .902,  .902}4.32 \\
		& MIPF & \textbf{0.54} & \textbf{0.54} & \textbf{0.57} & \textbf{0.59} & 3.00  & \textbf{1.86} & \textbf{2.09} & \textbf{2.18} \\
		& MINFFC & 0.58  & 0.59  & 0.60  & 0.62  & 4.00  & 3.41  & 3.27  & 3.00 \\
		\bottomrule
	\end{tabular}}
	\label{tab:mae}%
\end{table}%

In summary, the application of a noise filtering stage based on MIPF is beneficial for all the classifiers considered. 
In particular, the combination of MID and MIPF seems to be the most robust combination, showing the best accuracy values across all noise levels.
While some classifiers, as OSDL, are less affected by a previous preprocessing stage based on noise filtering, sensitive classifiers as MKNN take more advantage from noise filtering comparing No preprocessing against any other filtering technique. 
Nevertheless, MENN and MRNGE filters are not the best choices in noisy monotonic classification.
In the next section, we will try to get some insights on why MIPF is able to attain better performance than the compared algorithms and why MENN and MRNGE perform poorly.

\subsection{Monotonicity metrics}
\label{sec:metric_analysis}

Table \ref{tab:metrics} is dedicated to the monotonic metrics considered. The table is structured into five columns, first for the name of the algorithm and others for the noise levels studied. The results, grouped by metric, are the average metric values of the 12 datasets for the noise filtering algorithms, Relabelling and No preprocessing. The best result is stressed in bold.

\begin{table}[ht]
	\caption{Average of the monotonicity metrics with respect to monotonic noise filtering algorithms.}
	\label{tab:metrics}
	\centering
	\resizebox{.8\textwidth}{!}{
\begin{tabular}{c|l|cccc}
	\multicolumn{1}{r|}{} & Preprocessing & 0\% (Original) & 10\%  & 20\%  & 30\% \\
	\midrule
	\multirow{6}[2]{*}{NMI1} & No preprocessing & 0.021 & 0.024 & 0.026 & 0.028 \\
	& Relabelling & 0.002 & 0.001 & 0.001 & 0.002 \\
	& MENN & 0.005 & 0.007 & 0.008 & 0.009 \\
	& MRNGE & \textbf{0.001} & \textbf{0.001} & \textbf{0.000} & \textbf{0.000} \\
	& MIPF & 0.017 & 0.018 & 0.019 & 0.021 \\
	& MINFFC & 0.015 & 0.015 & 0.014 & 0.015 \\
	\midrule
	\multirow{6}[2]{*}{NMI2} & No preprocessing & 0.649 & 0.822 & 0.858 & 0.876 \\
	& Relabelling & 0.123 & 0.111 & 0.107 & 0.114 \\
	& MENN & 0.385 & 0.467 & 0.495 & 0.524 \\
	& MRNGE & \textbf{0.045} & \textbf{0.026} & \textbf{0.019} & \textbf{0.016} \\
	& MIPF & 0.368 & 0.405 & 0.455 & 0.487 \\
	& MINFFC & 0.263 & 0.257 & 0.245 & 0.263 \\
	\midrule
	\multirow{6}[2]{*}{Non-Comparable} & No preprocessing & 63265.24 & 67441.60 & 70829.14 & 74165.93 \\
	& Relabelling & 60444.83 & 71906.22 & 75223.07 & 69595.63 \\
	& MENN & 25806.34 & 10088.76 & 8691.59 & 8729.69 \\
	& MRNGE & \textbf{23132.81} & \textbf{8202.12} & \textbf{3830.44} & \textbf{3129.12} \\
	& MIPF & 38383.60 & 33834.46 & 30491.37 & 28302.81 \\
	& MINFFC & 29075.15 & 22584.23 & 17492.62 & 13228.82 \\
	\midrule
	\multirow{6}[2]{*}{Size} & No preprocessing & 657.41 & 657.41 & 657.41 & 657.41 \\
	& Relabelling & 657.41 & 657.41 & 657.41 & 657.41 \\
	& MENN & 372.39 & \textbf{277.89} & 257.80 & 249.65 \\
	& MRNGE & \textbf{335.69} & 278.98 & \textbf{245.15} & \textbf{229.25} \\
	& MIPF & 537.28 & 503.53 & 477.15 & 453.03 \\
	& MINFFC & 447.65 & 401.61 & 360.39 & 317.65 \\
	\bottomrule
\end{tabular}}
	
\end{table}

All  the metrics results in Table \ref{tab:metrics} are intrinsically related, but NMI1, NMI2 and \emph{Non-Comparable} are specific for the monotonic classification problem.
Observing \emph{Size}, the highest reduction rates corresponds to MRNGE. 
Average NMI1 and NMI2 indicate the grade of monotonicity in a data set: we must take as reference value NMI1 and NMI2 values for \emph{No preprocessing} at 0\% noise level. 
It is clear that in original data set the values are higher, while the monotonic noise removal techniques introduced in this paper reduce them.

\begin{figure}[t]
	\centering
	\begin{subfigure}[t]{.49\linewidth}
		\centering
		\includegraphics[width=1.0\textwidth]{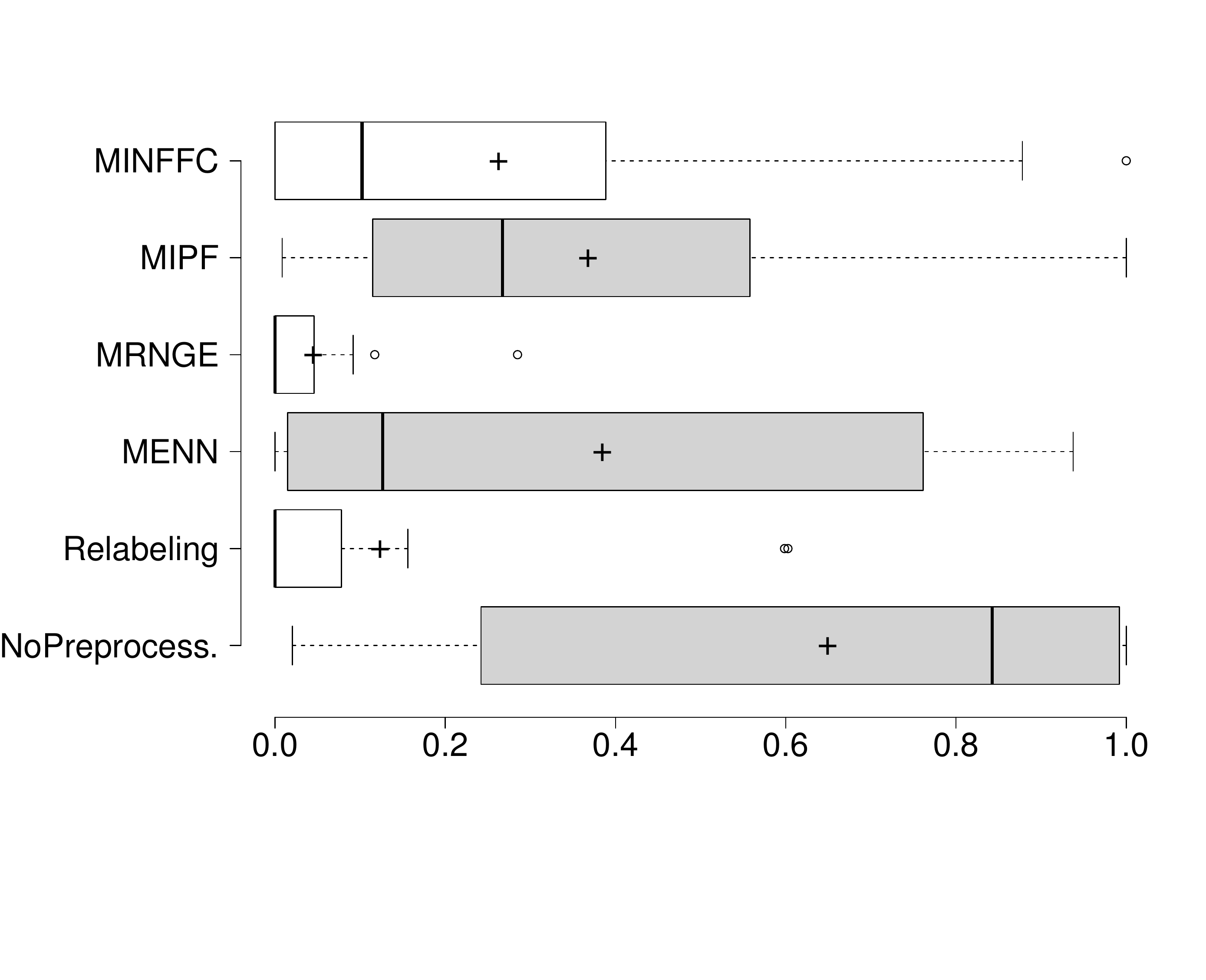}
		\caption{0\% noise}
		\label{fig:nmi2-0cn}
	\end{subfigure}
	\hfill
	\begin{subfigure}[t]{.49\linewidth}
		\centering
		\includegraphics[width=1.0\textwidth]{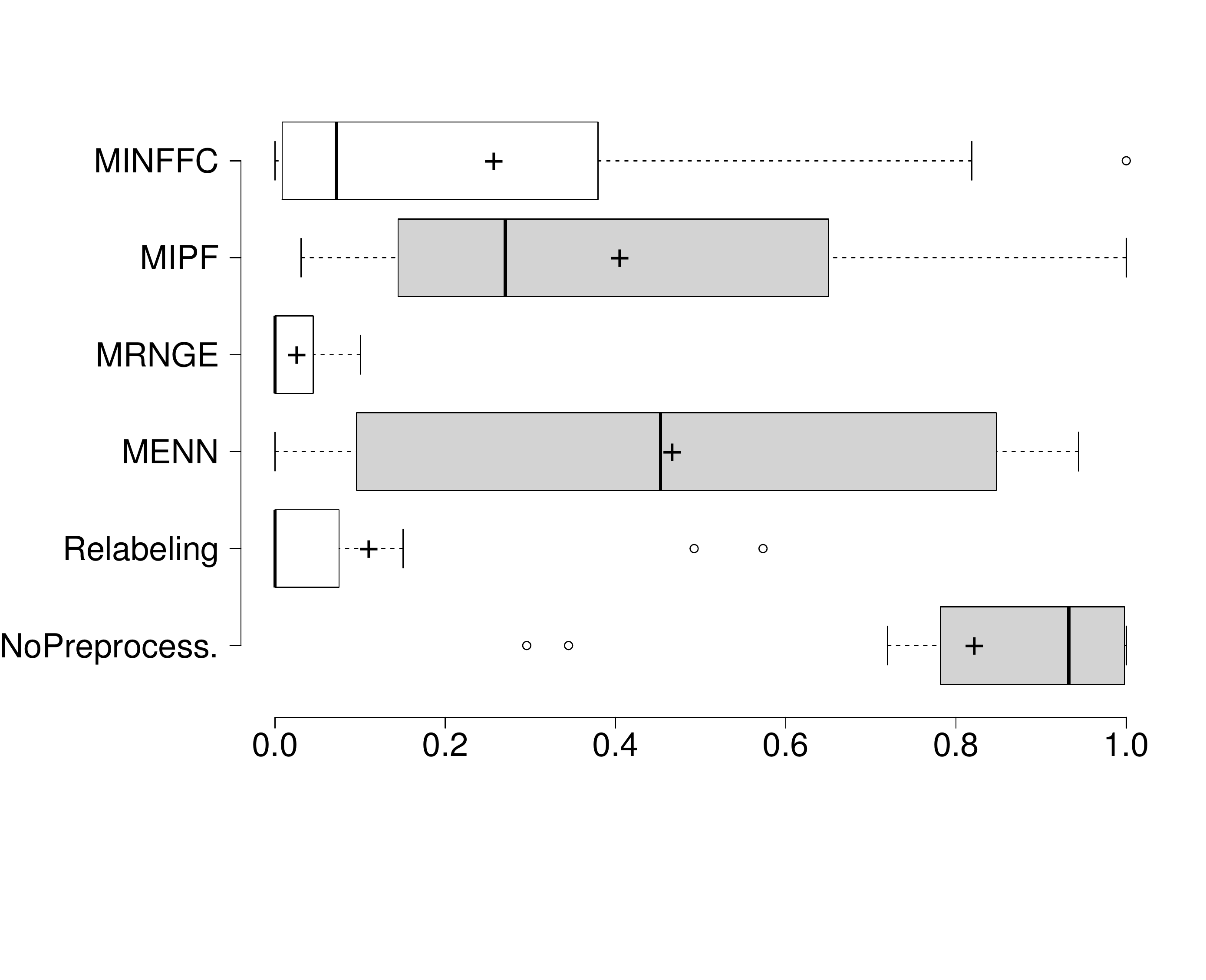}
		\caption{10\% noise}
		\label{fig:nmi2-10cn}
	\end{subfigure}
	\\
	\begin{subfigure}[t]{.49\linewidth}
		\centering
		\includegraphics[width=1.0\textwidth]{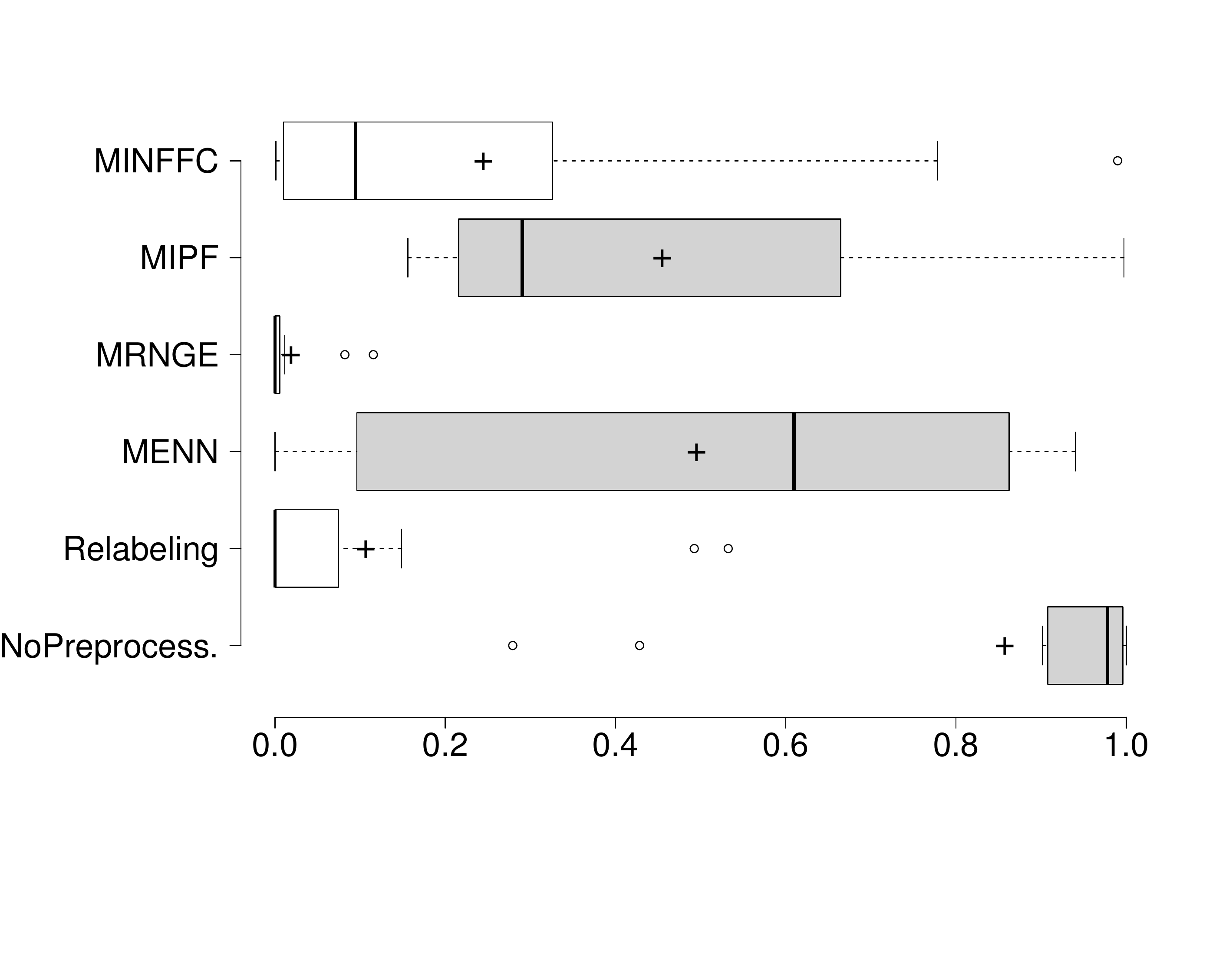}
		\caption{20\% noise}
		\label{fig:nmi2-20cn}
	\end{subfigure}
	\hfill
	\begin{subfigure}[t]{.49\linewidth}
		\centering
		\includegraphics[width=1.0\textwidth]{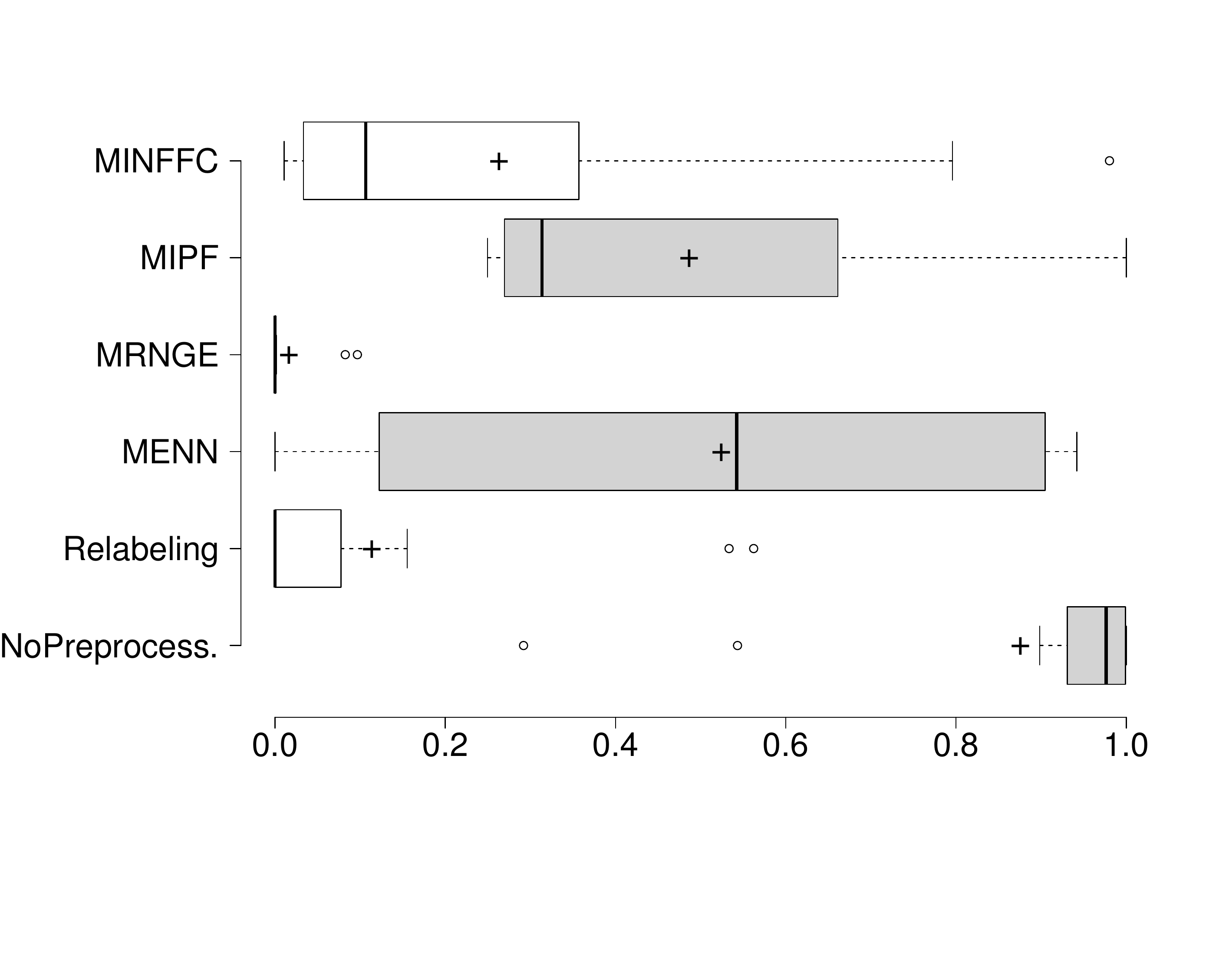}
		\caption{30\% noise}
		\label{fig:nmi2-30cn}
	\end{subfigure}
	\caption{NMI2 boxplots for each preprocessing technique in each noise level. Crosses indicate sample mean.}
	\label{fig:nmi2-boxplot}
\end{figure}

In Figure~\ref{fig:nmi2-boxplot} we present the boxplots for NMI2, as NMI1 shows very low variance and is much less descriptive.
As can be seen,  \emph{No preprocessing} decreases its variance as noise increases, while its median raises the noise introduced.
Relabelling achieves a stable behaviour, obtaining the same median and variance for all noise levels.
MENN  work reasonably well without noise, but adding more violations makes MENN perform much worse in terms of the variance shown.
Please note that MRNGE is also a similarity-based filter but, while MENN performs badly with noise, MRNGE achieves a low median and variance, which seems to be the best outcome so far.
However, MRNGE is not the best algorithm in terms of accuracy or MAE.
If we pay attention to MIPF and MINFFC, which obtain better performance than MRNGE with worse NMI2 values, their boxplots show an almost constant variance while the median slightly increases as noise does.
While the modification of the original data set by means of Relabelling aims to produce completely monotonic datasets with those rates equal to zero, it is interesting to note that MRNGE is able to reduce NMI1 and NMI2 even further than Relabelling. 

\begin{figure}[t]
	\centering
	\begin{subfigure}[t]{.49\linewidth}
		\centering
		\includegraphics[width=1.0\textwidth]{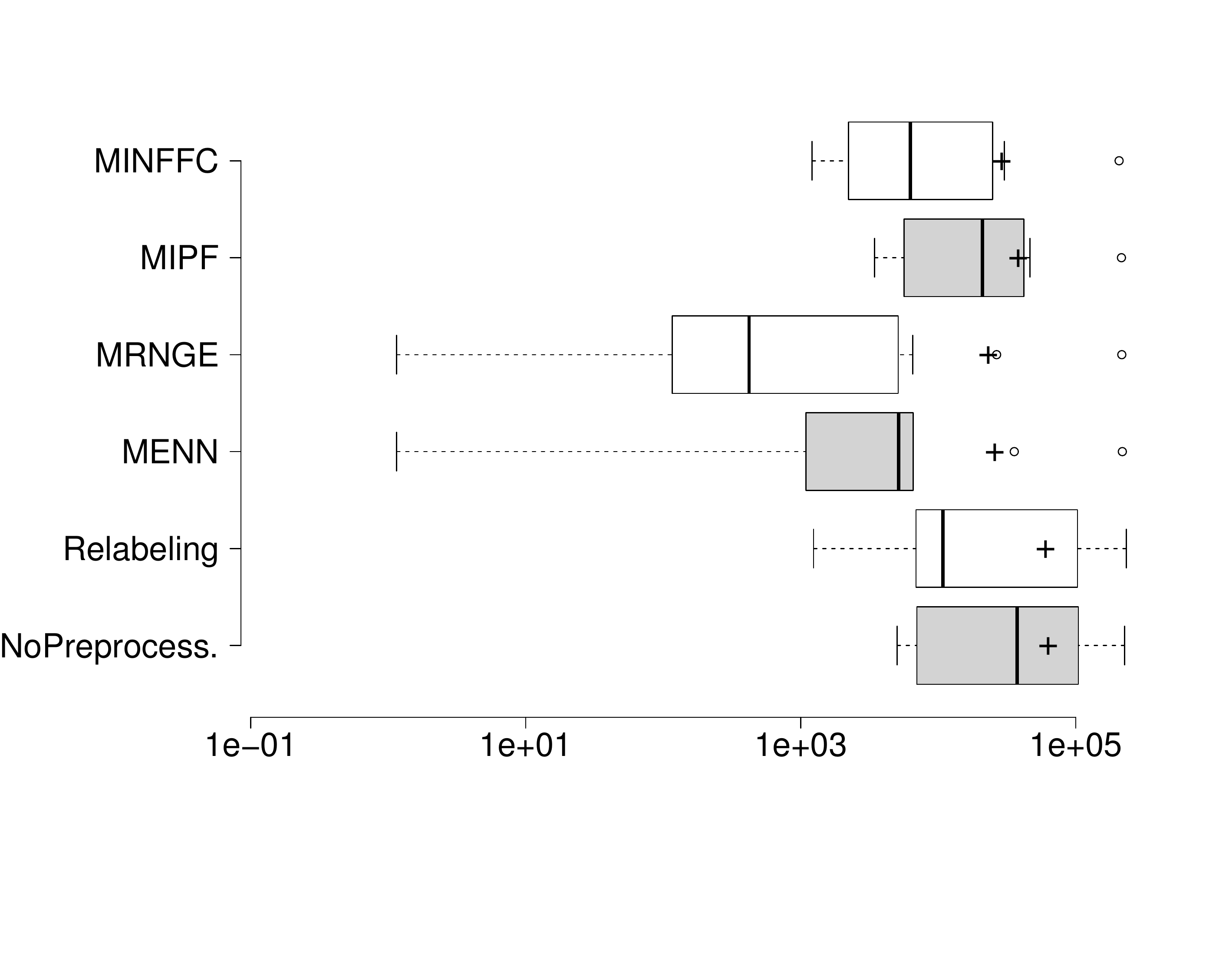}
		\caption{0\% noise}
		\label{fig:noncomparable-0cn}
	\end{subfigure}
	\hfill
	\begin{subfigure}[t]{.49\linewidth}
		\centering
		\includegraphics[width=1.0\textwidth]{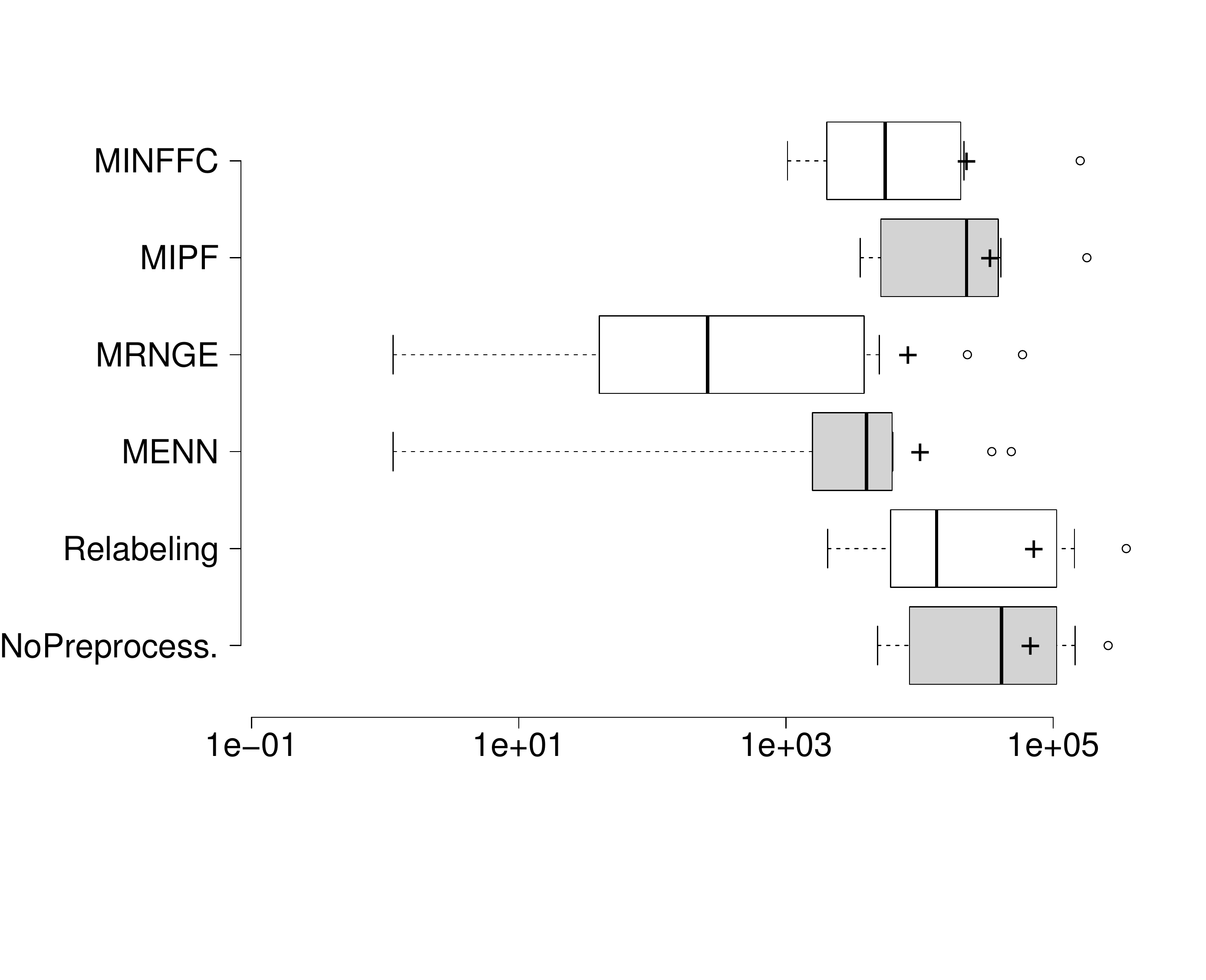}
		\caption{10\% noise}
		\label{fig:noncomparable-10cn}
	\end{subfigure}
	\\
	\begin{subfigure}[t]{.49\linewidth}
		\centering
		\includegraphics[width=1.0\textwidth]{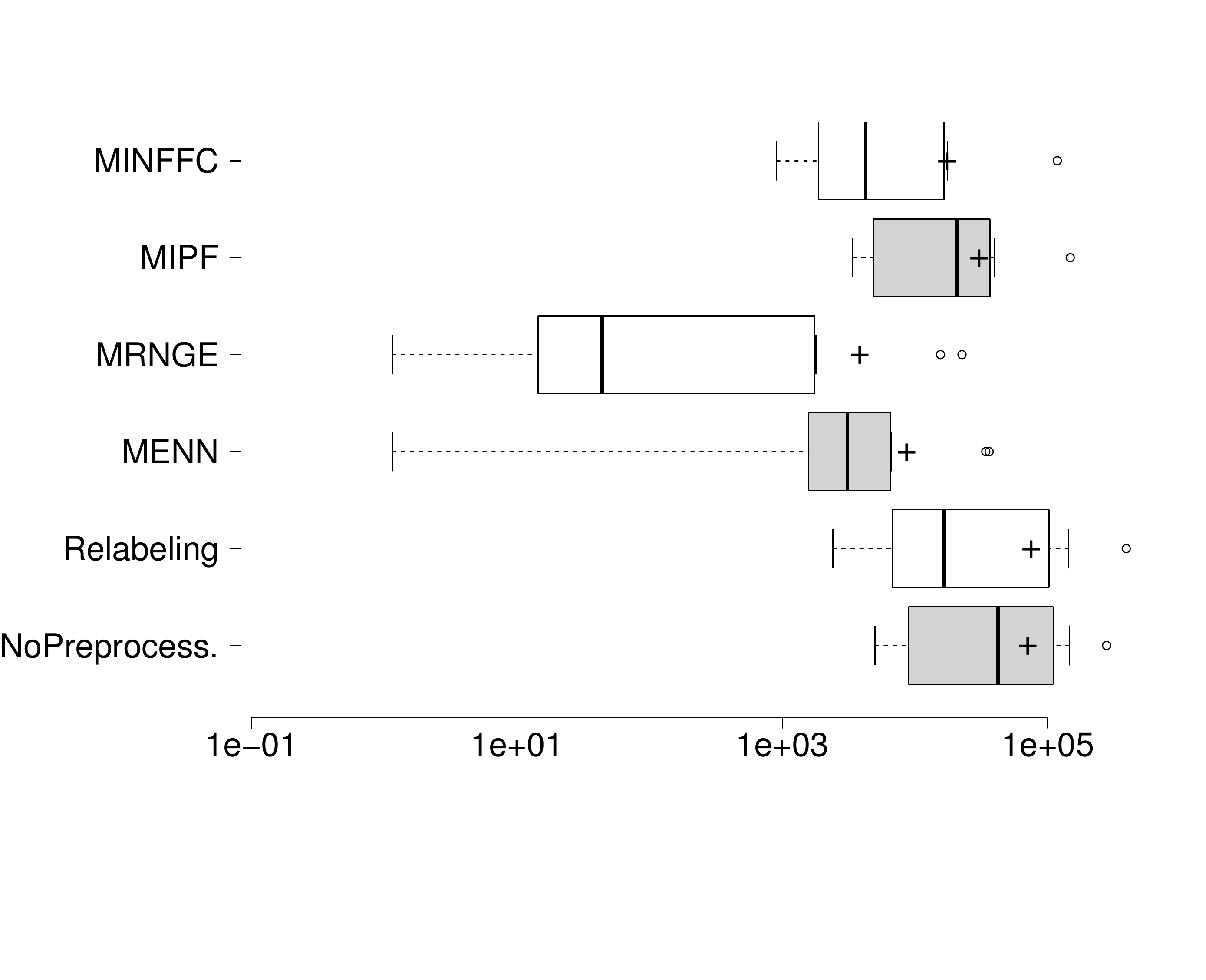}
		\caption{20\% noise}
		\label{fig:noncomparable-20cn}
	\end{subfigure}
	\hfill
	\begin{subfigure}[t]{.49\linewidth}
		\centering
		\includegraphics[width=1.0\textwidth]{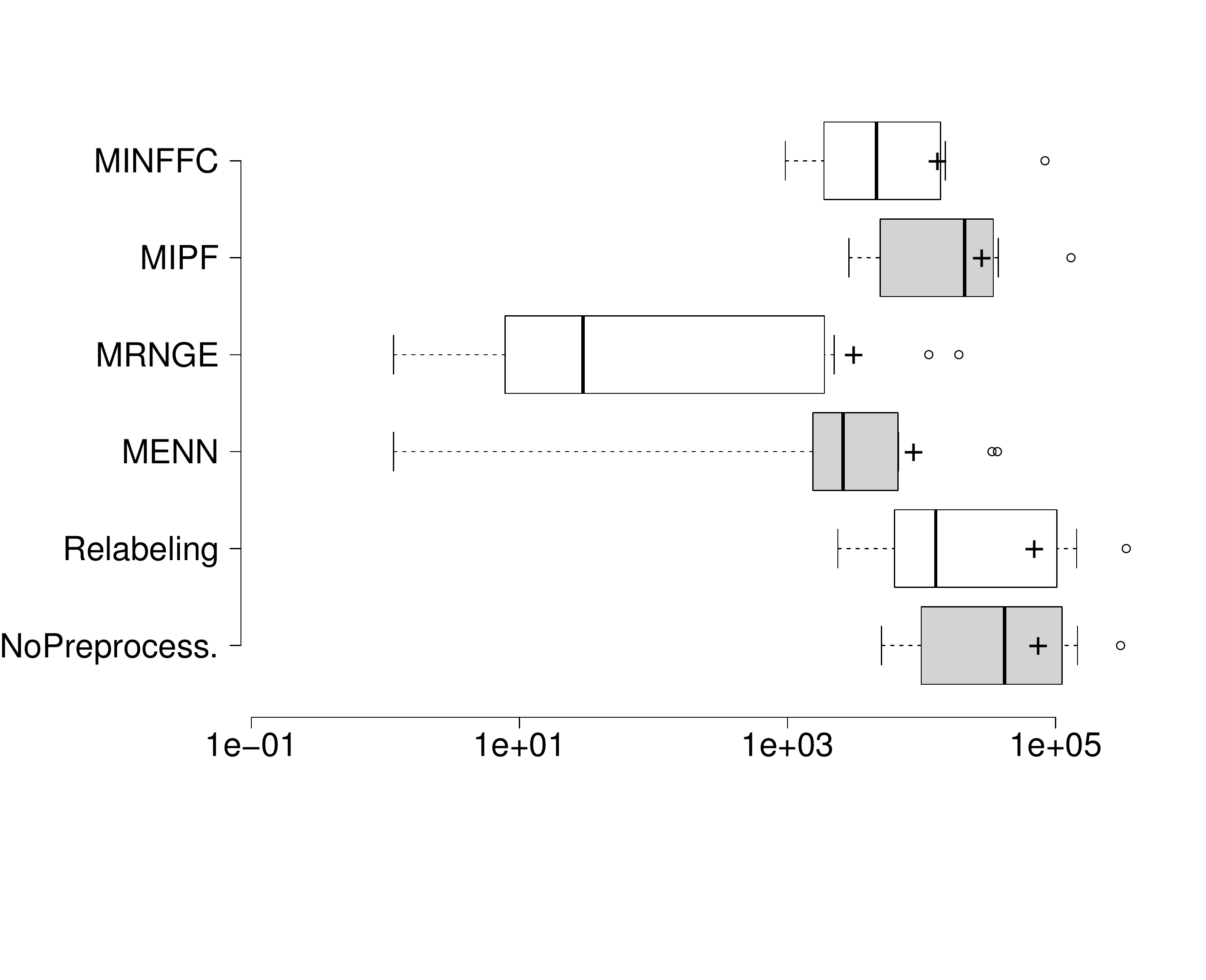}
		\caption{30\% noise}
		\label{fig:noncomparable-30cn}
	\end{subfigure}
	\caption{\emph{Non-comparable} boxplots for each preprocessing technique in each noise level.
	X axis is in log scale. Crosses indicate sample mean.}
	\label{fig:noncomparable-boxplot}
\end{figure}

We also want to pay attention to \emph{Non-comparable} values, as they indicate the number of monotonic violations that remain in the dataset after applying the different preprocessing techniques.
Figure~\ref{fig:noncomparable-boxplot} depicts the boxplots associated to the \emph{Non-comparable} values for each preprocessing approach.
We can observe that \emph{No preprocessing} and Relabelling show a similar variance but, while \emph{No preprocessing} has a higher median, Relabelling achieves a low median value.
The case of MENN and MRNGE is interesting, as they aim to reduce the number of violations as much as possible, and thus they achieve the best results for \emph{Non-comparable}. 
MIPF and MINFCC reduce their variance as the noise increases, but MINFFC is more exaggerated in this behaviour.
Since MIPF and MINFFC are the best performing algorithms, we may conclude that extreme behaviours as those shown by Relabelling or MRNGE are not desirable: while the former does not solve most of the violations, the latter tends to remove too many instances to eliminate the violations and altering the information contained in the dataset.

At this point, MRNGE is the preprocessing technique that is able to obtain the lowest amounts of non-comparable instances. 
However, we observed in Section~\ref{sec:performance_analysis} that MRNGE is not the best performing algorithm.
Since MRNGE also creates the most reduced datasets in terms of size, we may conclude that MRNGE is removing too many instances, which would lead to fewer violations of monotonic restrictions as shown by NMI1, NMI2 and \emph{Non-comparable} values.
This excessive removal will create an information loss in the dataset that penalizes the model obtained and thus showing poor performance in Accuracy and MAE values.

\begin{figure}[t]
	\centering
	\begin{subfigure}[t]{.49\linewidth}
		\centering
		\includegraphics[width=1.0\textwidth]{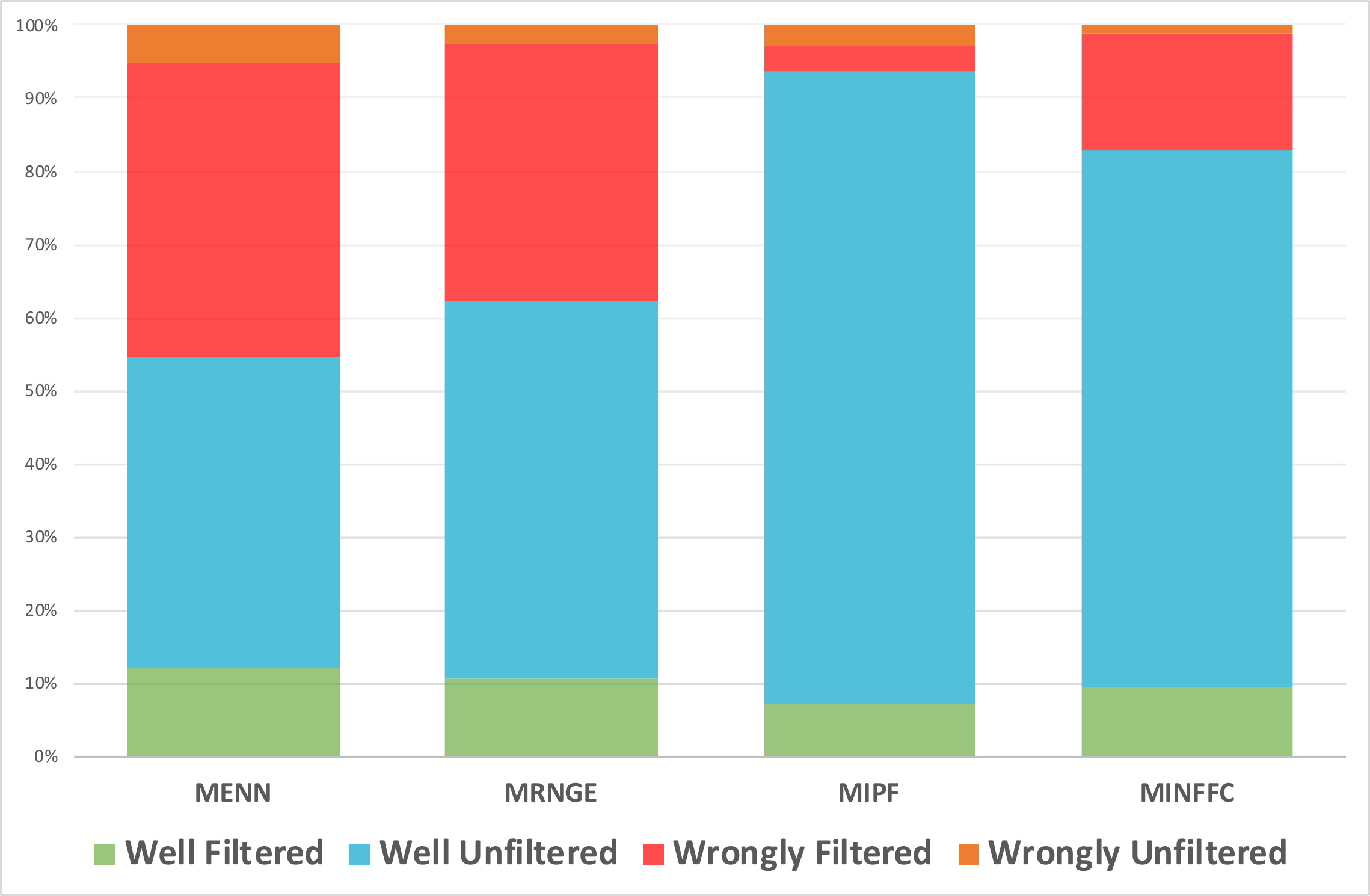}
		\caption{10\% noise}
		\label{fig:noiseStats-10cn}
	\end{subfigure}
	\\
	\begin{subfigure}[t]{.49\linewidth}
		\centering
		\includegraphics[width=1.0\textwidth]{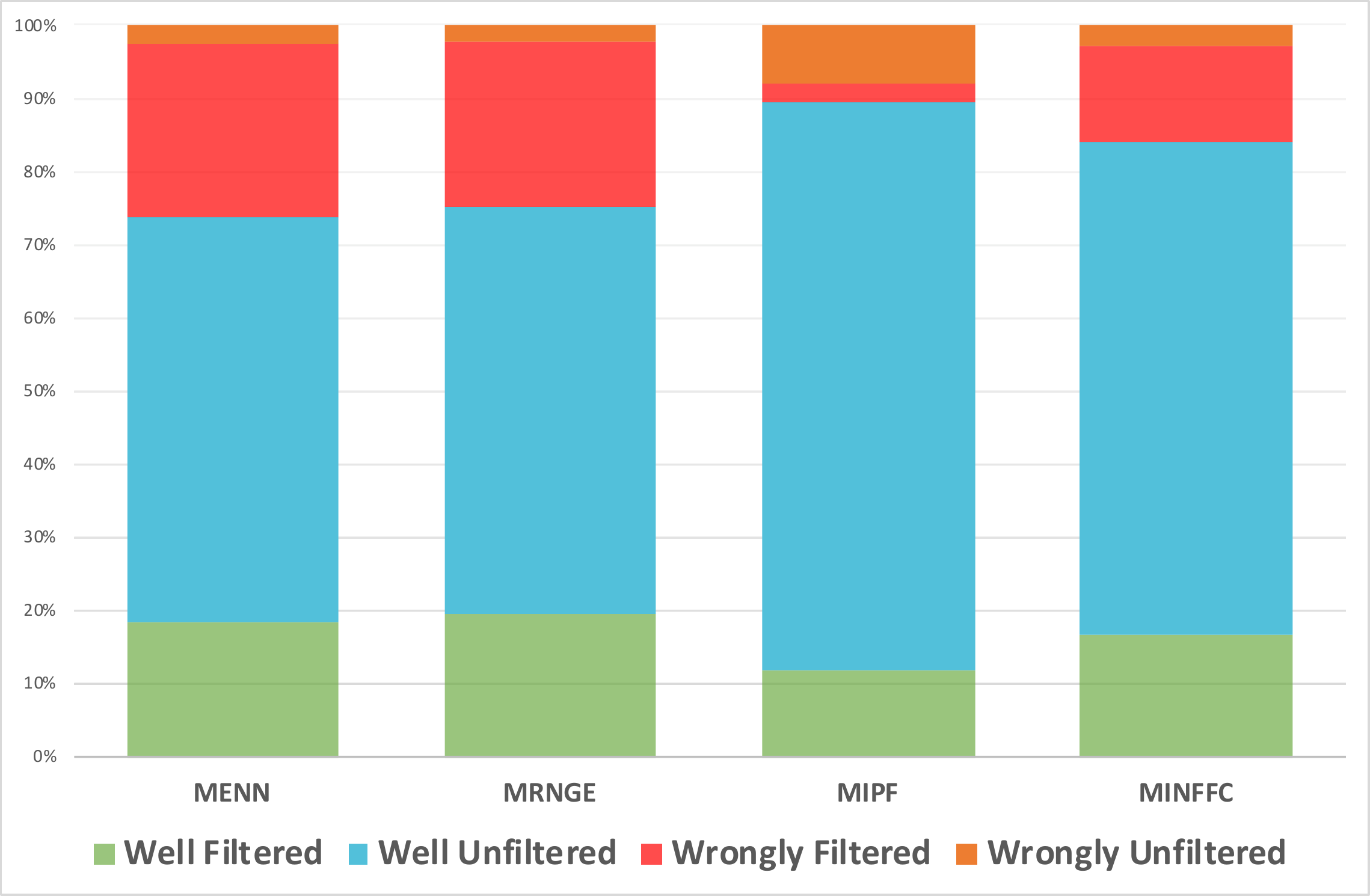}
		\caption{20\% noise}
		\label{fig:noiseStats-20cn}
	\end{subfigure}
	\hfill
	\begin{subfigure}[t]{.49\linewidth}
		\centering
		\includegraphics[width=1.0\textwidth]{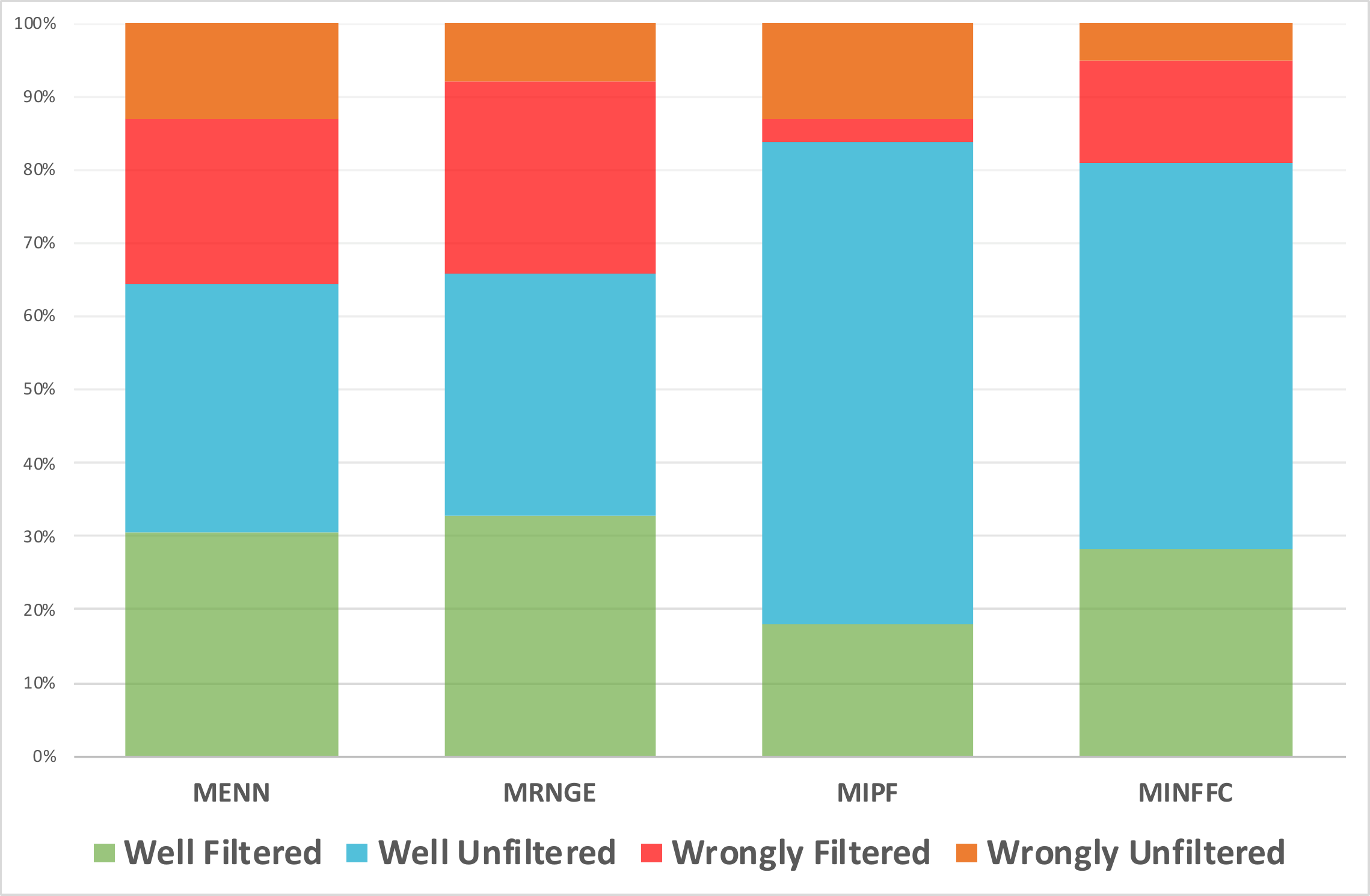}
		\caption{30\% noise}
		\label{fig:noiseStats-30cn}
	\end{subfigure}
	\caption{Percentages of the noise filters regarding to the successful/wrongly filtered instances for each noise level. Blue and green indicate correct decisions, while orange and red are related to wrong actions. The higher the sum of blue and green areas, the better.}
	\label{fig:noiseStats}
\end{figure}

An alternative way to analyze the behaviour of the different filters would be to examine how accurate is their noisy instances identification.
Figure~\ref{fig:noiseStats} shows the percentages of good and bad decisions of each noise filter, both in removing and keeping the instances in the dataset.
Since we need to know the corrupted instances to examine whether they were removed or not, we can only create these graphics for 10, 20 and 30\% noise levels.
As can be seen, MRNGE and MENN can eliminate all the noisy instances at 10\% noise level, but they also eliminate a large portion of the correct instances.
On the other hand, MIPF keeps some noisy instances in the dataset at 10\% noise level, but it is able to keep more correct instances than the others.

As the noise increases, MENN and MRNGE lose the ability to identify the correct examples, while they keep removing instances to nearly clear all the violations induced in the dataset.
At 30\% noise level, both MENN and MRNGE cannot keep enough correct instances in the dataset as MIPF does.
MINFFC is able to maintain a larger proportion of correct instances than MENN and MRNGE, but not as many as MIPF does.
The ability of MIPF of maintaining not noisy instances causes that it enables the classifiers to obtain the best results in accuracy and MAE, even when it has the largest amount of noisy instances compared to the other filtering algorithms.

We can observe that one big problem of some filters, MENN and MRNGE, is their over-filtering behaviour.
This problem was already detected in standard classification and motivated the proposal of advanced noise filters instead of just applying instance selection methods.
The leading solutions to this problem were the iterative removal of noise since cleaner datasets will help to posterior accurate filtering steps, and to establish a minimum amount of noise to be removed since no classifier can perfectly detect noisy instances and some false positives will always appear.
In this work we show that accurate monotonic classification filters also need to apply these mechanisms, as MIPF and MINFFC does, to avoid over-filtering in the dataset.

Finally, we must point out that the usage of monotonicity metrics alone cannot describe the ability of noise preprocessing algorithms in monotonic classification, as they can be largely minimized by removing too many instances as MRNGE does. 
Maintaining a good proportion of clean instances is crucial to enable the classifiers to obtain generalizable models. MIPF is the best approach analyzed in this respect.

\section{Experimental results and analysis on the benchmark data set: Winequality-red }
\label{sec:CasesOfStudy}

In this section, we apply the best combination filter-classifier  analyzed (MIPF + MID) in  Winequality-red, one of the benchmark datasets considered in the study in Section \ref{sec:experimentalFramework}. 
The goal is to analyze the effect of the filter process in the predictive models generated by the monotonic classifiers.

The Winequality-red data set was introduced in \cite{Cortez09} and can be found in the KEEL data set repository \citep{alcala2010keel}. It is related to a red variant of the Portuguese Vinho Verde wine. Due to privacy and logistic issues, only physicochemical (inputs) and sensory (the output) variables are available (e.g. there is no data about grape types, wine brand, wine selling price, etc.). 

The classes are ordered and not balanced (e.g. there are much more normal wines than excellent or poor ones).
 The data set consists of a sample of 1599 wines, described by 11 attributes and classified as 11 levels of quality. The attributes are:

\begin{itemize}
	\item Fixed Acidity (Fix), with values in the range [4.6,15.9].
	
	\item Volatile Acidity (Vol), with values in the range  [0.12,1.58].
	
	\item Citric Acid (Cit),  with values in the range [0.0,1.0].

	\item Residual Sugar (Res), with values in the range [0.9,15.5].
	
	\item Chlorides (Chl), with values in the range [0.012,0.611].
	\item Free Sulfur Dioxide (Fre), with values in the range  [1.0,72.0].
	\item Total Sulfur Dioxide (Tot), with values in the range [6.0,289.0]
	\item Density (Den), with values in the range [0.990,1.003].
	\item PH (Ph), with values in the range [2.74,4.01].
	\item Sulphates (Sul), with values in the range [0.33,2.0].
	\item Alcohol (Alc), with values in the range [8.4,14.9].
	
\end{itemize}

Real life datasets usually contain instances (wines in this case) which do not satisfy the monotonicity constraints. Table \ref{tab:breakingStud} presents two instances of the Winequality-red data set which produce a monotonicity collision between them. Lower values in features conclude with better qualification.

\begin{table}[!htp]
	\caption{Two instances which break the monotonicity constraint in Winequality-red data set.}
	\label{tab:breakingStud}
	\centering
	\resizebox{\linewidth}{!}{
	\begin{tabular}{cccccccccccc}\hline
		Fix & Vol & Cit & Res & Chl & Fre & Tot & Den & Ph & Sul & Alc &  Class \\		
		\hline
		
	     13.50 & 0.53 & 0.79 & 4.80 & 0.12 & 23.0 & 77.0 & 1.00 & 3.18 & 0.77 & 13.00 & 5 \\

	     11.20 & 0.28 & 0.56 & 1.90 & 0.07 & 17.0 & 60.0 & 0.99 & 3.16 & 0.58 & 9.80 & 6 \\
		
		\hline\end{tabular}	
	}
\end{table}

In this dataset, there are 471 instances with one or more monotonic collisions among them, which significantly affects to the monotonicity of the prediction models.

The filter methods are necessary to reduce these number of collisions, thus the generated model is able to keep its prediction capabilities while the monotonicity restrictions are taken into account. 

To analyze this situation, the data set has been evaluated following a 10-fcv using MID, and MIPF+MID. The average performance they offered appears in Table \ref{tab:WineqRes}. As the analysis in the previous section reflects, the best prediction models are produced by the MID algorithm, while its combination with MIPF keeps similar prediction capabilities improving all the monotonicity metrics, and reducing the  size of the model (number of rules).

\begin{table}[!htp]
	\caption{Results in Winequality-red data set for MID and MIPF+MID.}
	\label{tab:WineqRes}
	\centering
	\resizebox{\linewidth}{!}{
	\begin{tabular}{ccccccc}\hline
			& 	NMI1 & NMI2 &  Non-Comp & Size & MAE &  \#Branches \\		
		\hline
		
		MID &  0.00032 & 0.31345 & 331872 & 1439.1 & 0.46216 & 344 \\
		
		MIPF+MID & 0.00030 & 0.29717 & 311433.75 & 1404.4 & 0.47275 &  321\\
		
		\hline\end{tabular}	
	}
\end{table}

\begin{figure}[tp]
	\centering
	\begin{subfigure}[t]{.49\linewidth}
		\centering
		\includegraphics[width=1.0\textwidth]{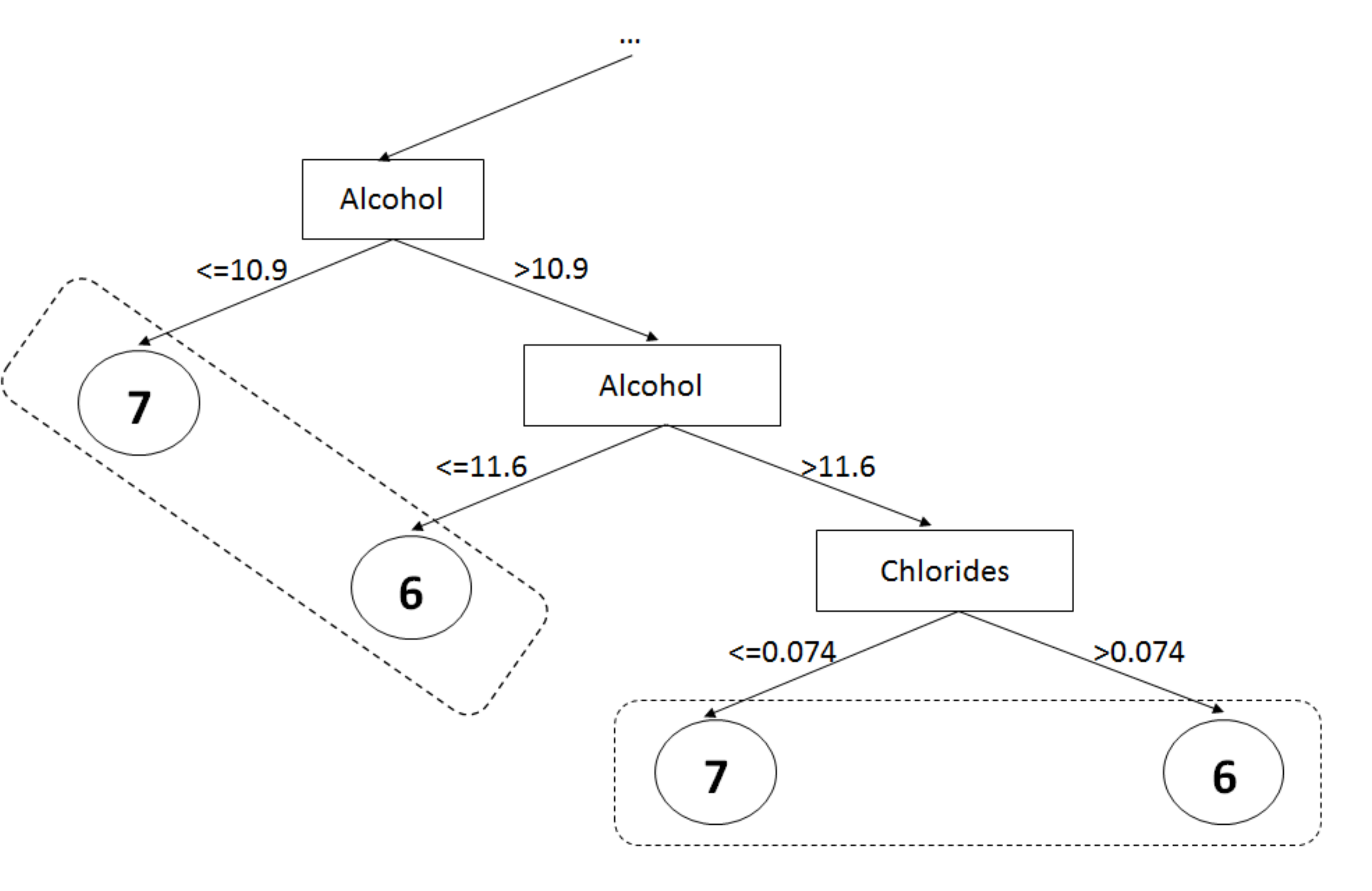}
		\caption{MID without MIPF}
		\label{fig:figMID}
	\end{subfigure}
	\hfill
	\begin{subfigure}[t]{.49\linewidth}
		\centering
		\includegraphics[width=1.0\textwidth]{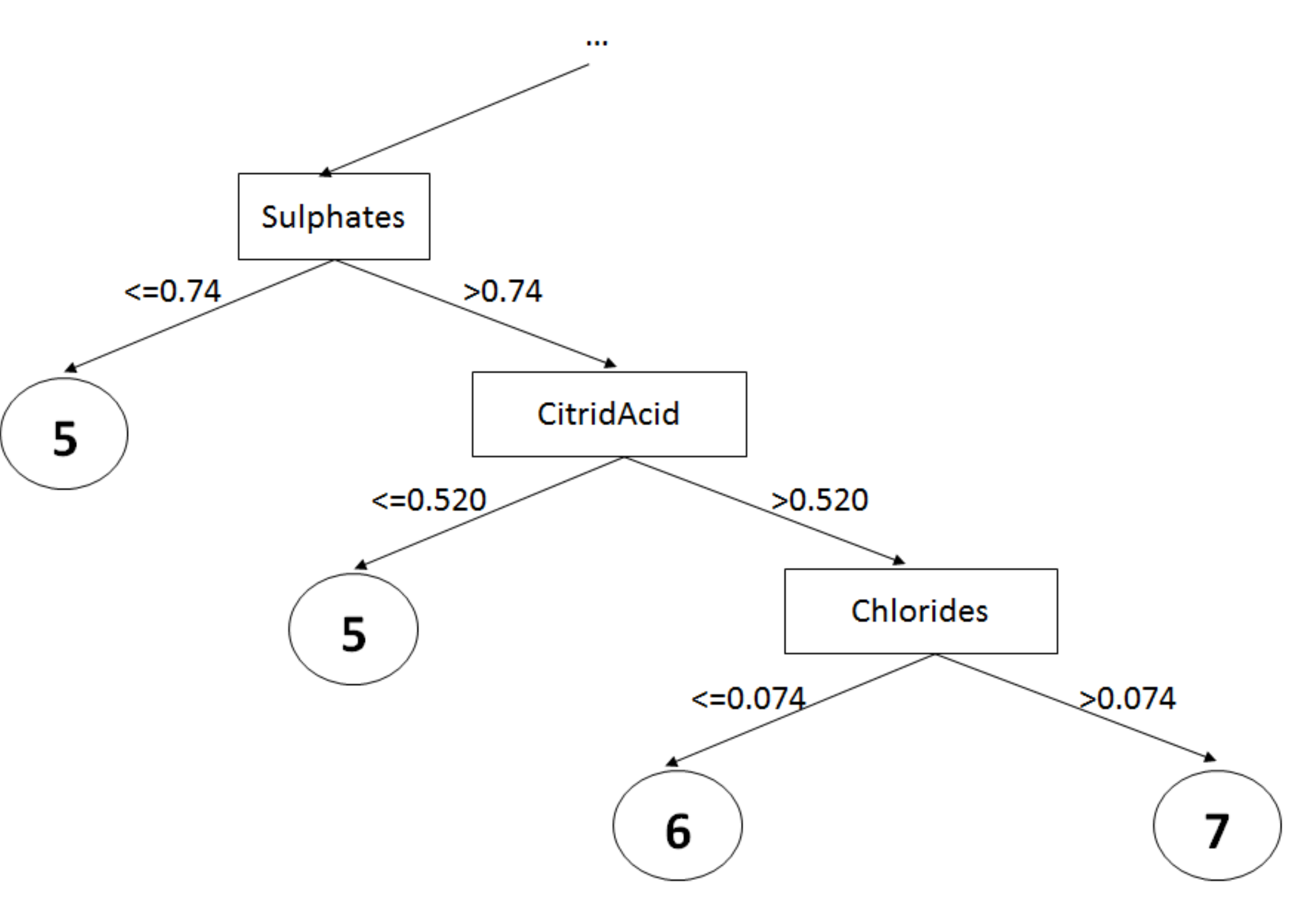}
		\caption{MID with MIPF}
		\label{fig:figMIPFMID}
	\end{subfigure}
	\caption{Decision trees extracted from Winequality-Red using MID with and without MIPF (dotted boxes reflect violations of the monotonicity constraints).}
	\label{fig:decisionTrees}
\end{figure}

The removed instances produce many differences in the decision trees generated by the MID algorithm, as we can see in Figure  \ref{fig:decisionTrees} where we plot a portion of the tree extracted from the same partition of Winequality-red data set by MID, without and with MIPF filter. The dotted boxes in Subfigure \ref{fig:figMID} reflect violations of the monotonicity constraints.  As it can be seen in Table \ref{tab:WineqRes} and Figure \ref{fig:decisionTrees}, the structure of the decision tree changes significantly in the number of branches (last column in Table \ref{tab:WineqRes}), antecedent order and partition range, which transforms the meaning of the rules. In addition, the filter method is able to resolve many of the monotonicity conflicts which already appear using the MID classifier isolate.

\section{Conclusions}
\label{sec:Conclusions}
In this paper, we have proposed the use of noise filtering algorithms as a preprocessing stage to decrease the monotonicity violations present in the original data. 
We have analyzed four noise removal algorithms, adapted to the monotonic domain, using different prediction rates and metrics over a high number of datasets, coming from standard classification and regression problems. 
The main conclusions related to the analyzed algorithms are:

\begin{itemize}
	\item Monotonic noise removal algorithms are able to remove instances which negatively affect to the monotonicity of real data, altering the lowest possible the concepts represented in the original data and improving the efficiency and efficacy of the monotone classifiers.
	
	\item Relabelling is not able to deal with noisy environments, as its premises are skewed by the corrupted instances, wrongly relabelling instances and creating data shifting in the training set with respect to the test partition. 
	
	\item  Monotonic Iterative Partitioning Filtering (MIPF) is able to preserve and even to improve the prediction performances offered by classical monotonic classifiers such as   M$k$NN, OLM, OSDL and MID. 
	
	\item In the particular case of Winequality-red analyzed as an example, MIPF affects positively to the monotonicity constraints associated with the models extracted by MID.

\end{itemize}

We have also shown that monotonicity metrics cannot describe what constitutes a good filtering, as they can be biased by removing too many instances.
While we have shown that filtering can greatly help to diminish the impact of noisy instances in monotonic classification, there is still promising options to explore: although Relabelling is not designed to work with noisy instances, correct reparation of an instance can greatly help to improve even further the results of this work. 
Since the monotonicity metrics can deceive the noise filters, other measures can be designed to avoid the greedy removal of preprocessing techniques.

\section*{Acknowledgements}
This work was supported by TIN2014-57251-P and TIN2017-89517-P, by the Spanish ''Ministerio de Economía y Competitividad'' and by ''Fondo Europeo de Desarrollo Regional'' (FEDER) under Project TEC2015-69496-R and the Project BigDaP-TOOLS - Ayudas Fundaci\'on BBVA a Equipos de Investigaci\'on Cient\'{\i}fica 2016.

\section*{References}
\bibliography{refs}   

\end{document}